\documentclass[10pt,twocolumn,letterpaper]{article}
\usepackage{wacv}              

\usepackage[accsupp]{axessibility} 

\usepackage{graphicx}
\usepackage{amsmath}
\usepackage{amssymb}
\usepackage{booktabs}

\usepackage[pagebackref,breaklinks,colorlinks]{hyperref}
\usepackage[capitalize]{cleveref}
\crefname{section}{Sec.}{Secs.}
\Crefname{section}{Section}{Sections}
\Crefname{table}{Table}{Tables}
\crefname{table}{Tab.}{Tabs.}
\def\wacvPaperID{*****} 
\def\confName{WACV}
\def\confYear{2025}

\makeatletter
\renewcommand\paragraph{\@startsection {paragraph}
{4}
{\z@}
{1.5ex \@plus .3ex \@minus .2ex}
{-2ex \@plus -.5ex \@minus -2ex}
{\normalfont\bfseries\raggedright}}
\makeatother

\usepackage[utf8]{inputenc} 
\usepackage[T1]{fontenc}    
\usepackage{url}            
\usepackage{booktabs}       
\usepackage{amsfonts}       
\usepackage{nicefrac}       
\usepackage{microtype}      
\usepackage{xcolor}         

\usepackage{amsmath}
\usepackage{amssymb}
\usepackage{amsthm}
\usepackage{gensymb}
\usepackage{arydshln}
\usepackage{multicol}
\usepackage{float}
\usepackage{mathtools}

\usepackage[ruled,noline,noend]{algorithm2e}
\SetArgSty{textnormal}
\SetAlFnt{\footnotesize}

\DontPrintSemicolon

\newcommand{\depthmetric}[1]{\ensuremath{\delta_{\leq 1.25}}}

\newcommand\cut[1]{}

\newif\ifauthorstmtonly
\authorstmtonlyfalse

\newif\ifsupplementaryonly
\supplementaryonlyfalse

\ifauthorstmtonly\else
\title{SANPO \\
A Scene Understanding, Accessibility and Human Navigation Dataset}
\ifsupplementaryonly\else
\usepackage[para]{footmisc}

\author{Sagar M. Waghmare\thanks{Google Research}\\
\and
Kimberly Wilber\footnotemark[1] \thanks{Work done while at Google}\\
\and
Dave Hawkey\footnotemark[1]\\
\and
Xuan Yang\footnotemark[1]\\
\and
Matthew Wilson\footnotemark[1]\\
\and
Stephanie Debats\footnotemark[1]\\
\and
Cattalyya Nuengsigkapian\footnotemark[1]\\
\and
Astuti Sharma\footnotemark[1]\\
\and
Lars Pandikow\thanks{Parallel Domain}\\
\and
Huisheng Wang\footnotemark[1]\\
\and
Hartwig Adam\footnotemark[1]\\
\and
Mikhail Sirotenko\footnotemark[1]
}
\fi
\fi

\begin{document}

\ifauthorstmtonly
\else

  \ifsupplementaryonly

    \appendix
\section{Appendix}

\subsection{Data Collection}
\subsubsection{Rig}
\label{appendix:rig}
To accommodate SANPO-Real's multi stereo camera requirements we designed a specialized data collection rig. This rig prioritizes hardware integration, reliable GPU cooling, and comfort for the wearer. Our setup involves volunteers wearing head and chest mounted ZED cameras (ZED-M and ZED-2i, respectively), with supporting hardware in a backpack. We also developed a mobile app for visualization and to control the data collection. Figure~\ref{fig:rig} shows the data collection system in action.
\begin{figure}[h!]%
    \centering
    \subfloat{{\includegraphics[width=0.15\linewidth]{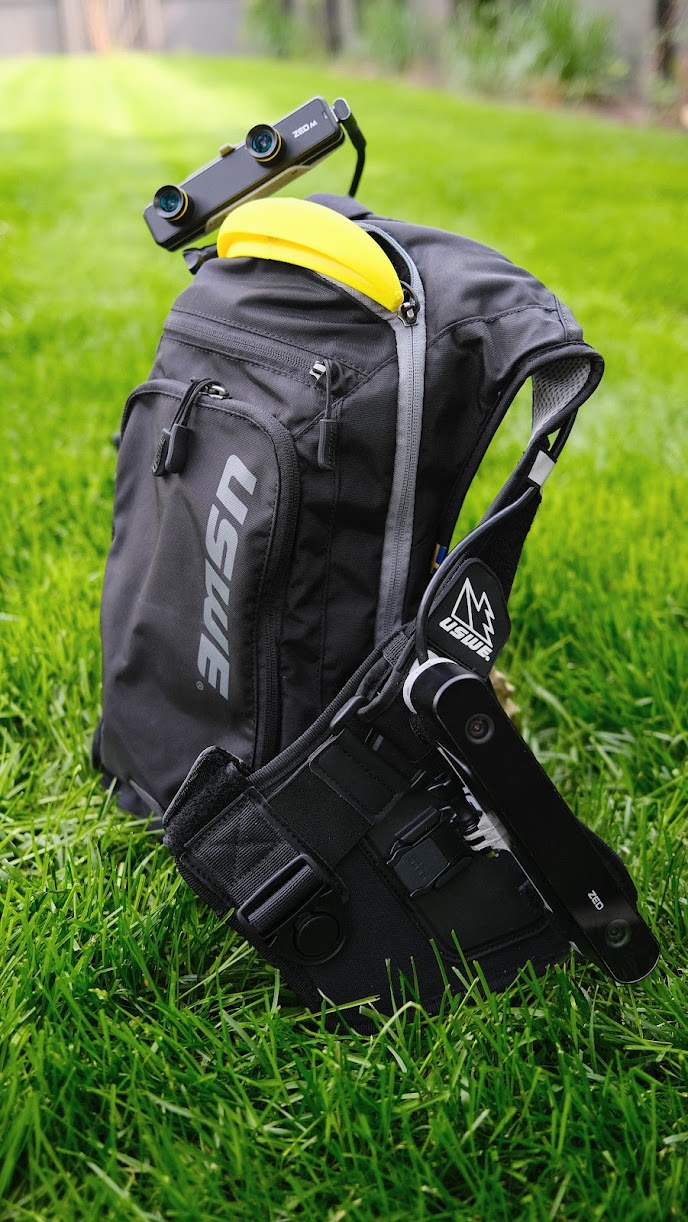}}}%
    \hspace{1pt}
    \subfloat{{\includegraphics[width=0.15\linewidth]{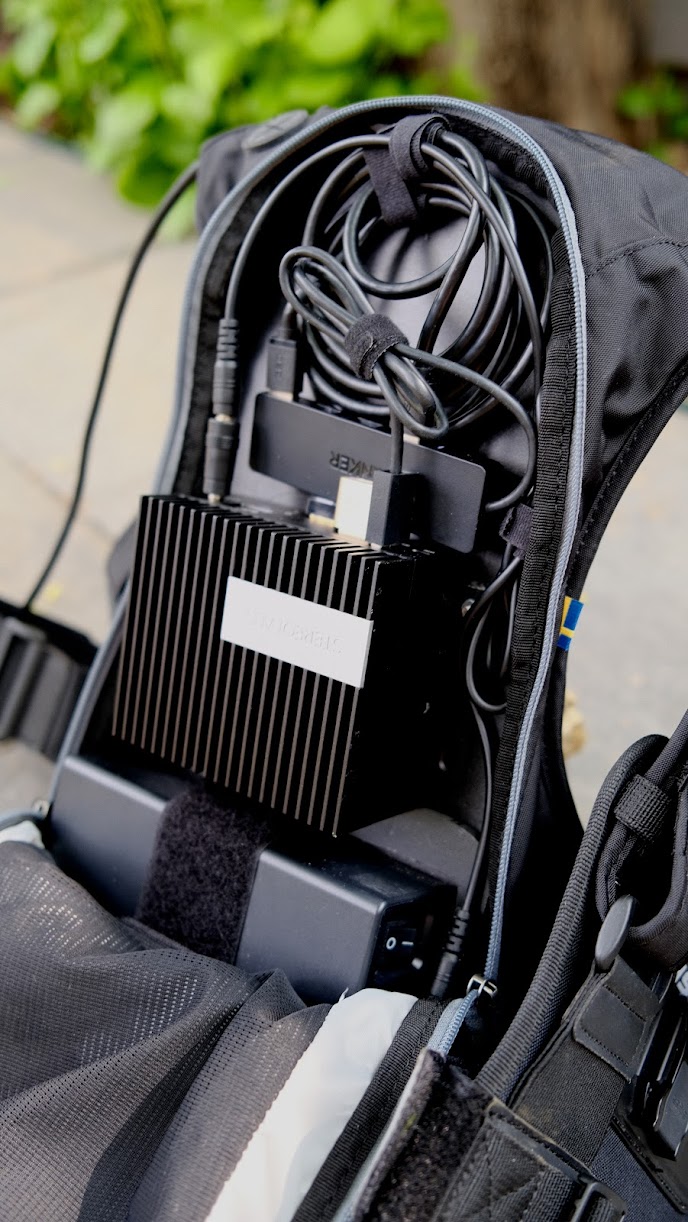}}}%
    \hspace{1pt}
    \subfloat{{\includegraphics[width=0.15\linewidth]{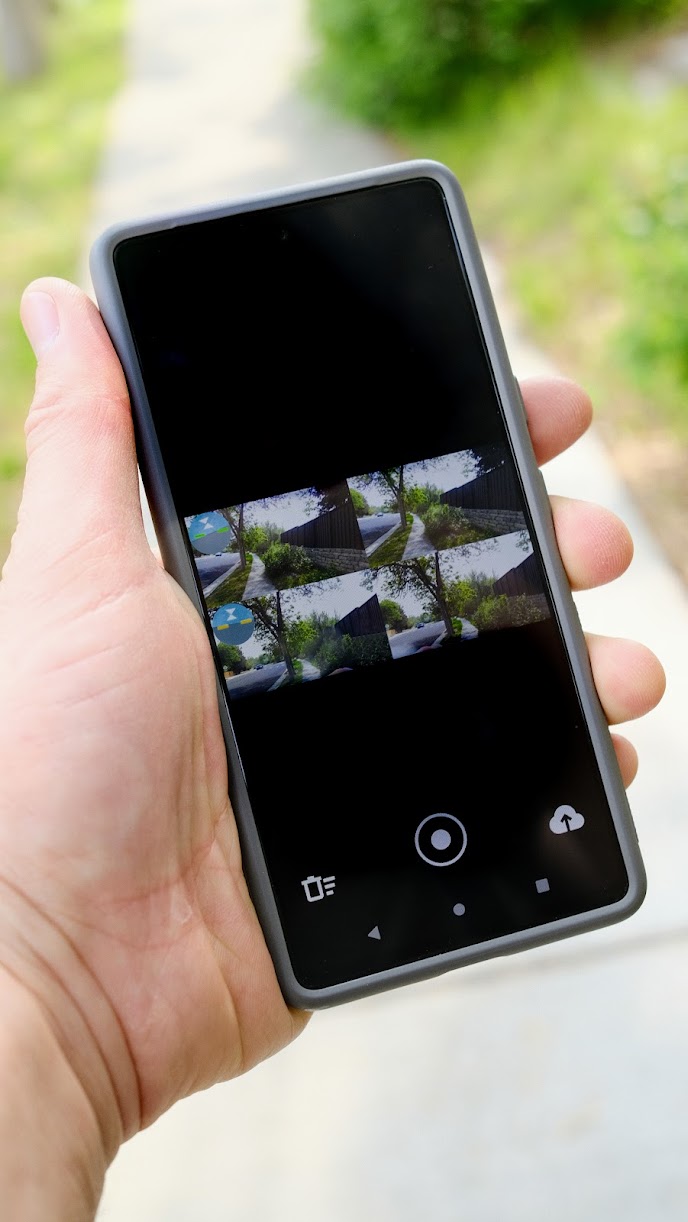}}}%
    \hspace{1pt}
    \subfloat{{\includegraphics[width=0.2\linewidth, height=0.2665\linewidth]{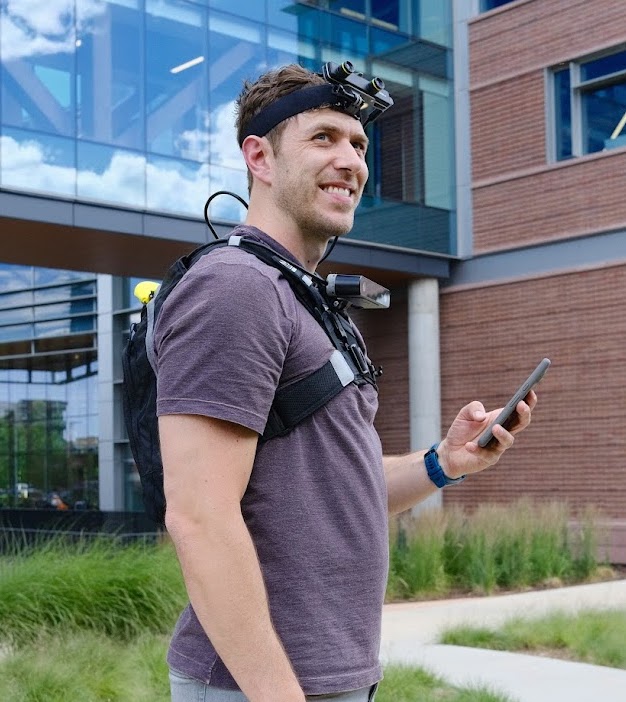}}}%
    
    \caption{\textbf{SANPO-Real Data Collection Rig.}}%
    \label{fig:rig}%
\end{figure}

\subsection{Dataset}

\subsubsection{SANPO-Synthetic Reproducibility and rendering environment.}
\label{appendix:sanpo-synthetic-reproducibility}
We created SANPO-Synthetic through our collaboration with a third party, Parallel Domain. If other researchers wish to reproduce these environments with other tools (NVidia Omniverse, Unreal, Unity, etc), we would welcome that. To aid reproducibility, here are detailed specifications for SANPO-Synthetic's virtual rendering environment.

All \% are at session level.
\begin{enumerate}

    \item Scene types : Urban environments only.

    \item Camera Type : Zed 2i
    \begin{enumerate}
        \item Image width: 2208
        \item Image height: 1242
        \item fx: 1914.203
        \item fy: 1914.203
        \item cx: 1074.4403
        \item cy: 655.79846
        \item camera matrix: \\ 
        $$\begin{pmatrix*}[r]
            1914.203 & 0 & 1074.4403 \\
            0 & 1914.203 & 655.79846 \\
            0 & 0 & 1        
        \end{pmatrix*}$$

        \item stereo transform (between left/right cameras): \\
        $$\begin{pmatrix*}[r]
            1 & 0 & 0 & 119.96817 \\
            0 & 1 & 0 & 0 \\
            0 & 0 & 1 & 0 \\
            0 & 0 & 0 & 1
        \end{pmatrix*}$$
    \end{enumerate}

    \item Camera Type : Zed Mini
    \begin{enumerate}
        \item image\_width: 2208
        \item image\_height: 1242
        \item fx: 1376.4702
        \item fy: 1376.4702
        \item cx: 1112.7797
        \item cy: 599.8397
        \item camera matrix: \\
        $$\begin{pmatrix*}[r]
            1376.4702 & 0 & 1112.7797 \\
            0 & 1376.4702 & 599.8397 \\
            0 & 0 & 1 
        \end{pmatrix*}$$
        \item stereo transform (between left/right cameras): \\
        $$\begin{pmatrix*}[r]
            1 & 0 & 0 & 62.944813 \\
            0 & 1 & 0 & 0 \\
            0 & 0 & 1 & 0 \\
            0 & 0 & 0 & 1
        \end{pmatrix*}$$
    \end{enumerate}
 
    \item Camera Positions
    \begin{enumerate}
        \item Zed 2i on chest.
        \item Zed mini just above head.
        \item Both with natural tilt variations.
        \item 50\% of sessions from each position. 
    \end{enumerate}
    
    \item FPS (Frames per second)
    \begin{enumerate}
        \item 60\% at 5 FPS.
        \item 20\% at 14.28 FPS.
        \item 20\% at 33.33 FPS.
    \end{enumerate}
    
    \item Ground truth annotations
    \begin{enumerate}
        \item Panoptic segmentation mask.
        \item Metric depth map.
    \end{enumerate}

    \item Lighting and Weather
    \begin{enumerate}
        \item 70\% well lit sunny 
        \item 10\% are at dawn/dusk with the sun low in the horizon
        \item 10\% are dark/nighttime
        \item 5\% have fog 
        \item 5\% have rain
    \end{enumerate}

    \item Obstacles
    \begin{enumerate}

        \item Garbage can
        \begin{enumerate}
            \item 50\% One per street block.
            \item 30\% two garbage cans. E.g: One normal and one recycle.
            \item 20\% no garbage can.
        \end{enumerate}
        
        \item Trash bags
        \begin{enumerate}
            \item 50\% None
            \item 40\% 1-2
            \item 10\% >=5
        \end{enumerate}
        
        \item Bike racks : One per street block.
        
        \item Mailbox
        \begin{enumerate}
            \item 60\% one per street block.
            \item 20\% two adjacent mailboxes per street block.
            \item 20\% None.
        \end{enumerate}

        \item Fire Hydrant
        \begin{enumerate}
            \item 80\% One per street block.
            \item 20\% None.
        \end{enumerate}

        \item Construction cones : As provided by the rendering scene map.
    \end{enumerate}
    
    \item Road Vehicle : Low, mid and high is the setting in the rendering engine.
    \begin{enumerate}
        \item 20\% None
        \item 30\% low 
        \item 30\% mid
        \item 20\% high
    \end{enumerate}
    
    \item Pedestrians
    \begin{enumerate}
        \item 10>= per street block (50\%) 
        \item 5>= per street block (30\%)
        \item <3 per street block (20\%)
        \item 20\% very close to the ego person.
    \end{enumerate}
    
    \item Trees on sidewalk
    \begin{enumerate}
        \item 60\% high density.
        \item 20\% low density.
        \item 20\% no trees on the sidewalk.
    \end{enumerate}

    \item Other naturally occurring things like curbs, dips, crosswalks, parking meters, traffic signs and lights, fences, plants, hedges etc.. will be included as provided by the rendering scene map.
    
    \item Not Supported
    \begin{enumerate}
        \item Bike paths.
        \item Riders on sidewalk.
        \item Foliage and seasonal color changes of leaves.
    \end{enumerate}

\end{enumerate}

\subsubsection{Dataset Comparison}
\label{appendix:dataset_comparison}
\begin{table*}
\newcommand{\icon}[1]{{\includegraphics[bb=0 0 2cm 2cm, height=10pt,viewport={0pt 120pt 480pt 430pt}]{#1}}}
\newcommand{\smicon}[1]{{\includegraphics[bb=0 0 2cm 2cm,height=10pt,viewport={0pt 120pt 480pt 500pt}]{#1}}}
\newcommand{\emptystar}[0]{\icon{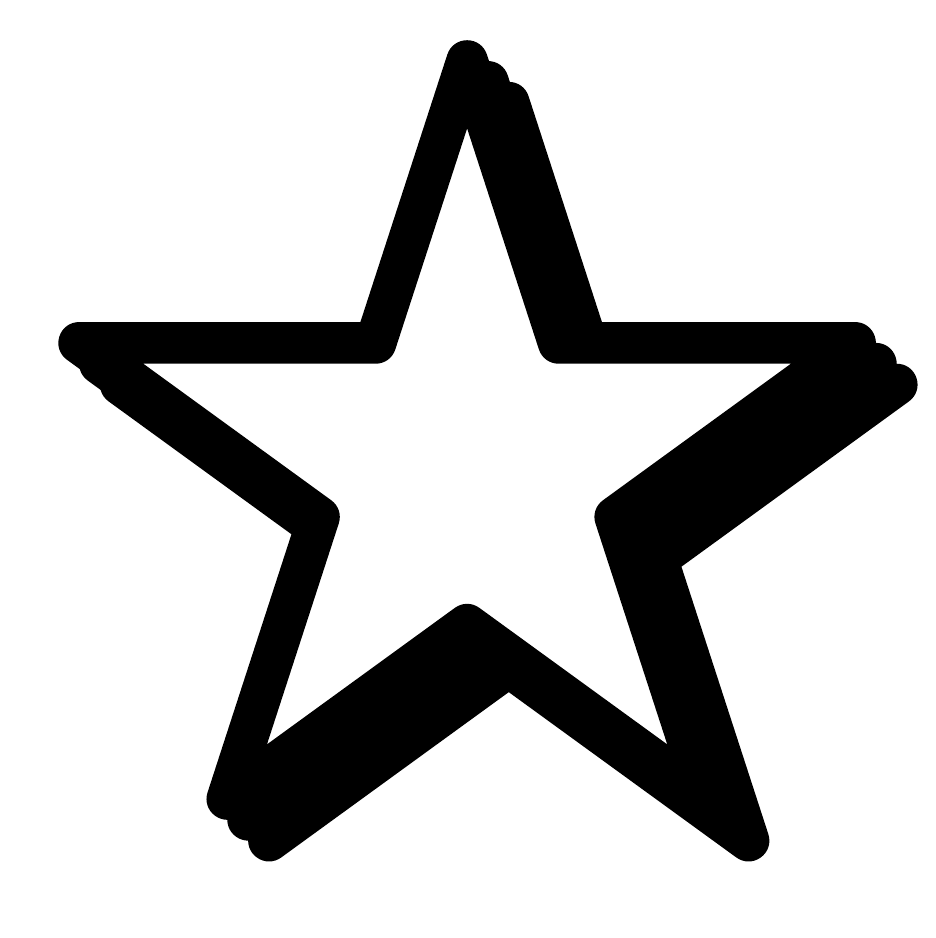}}
\newcommand{\smemptystar}[0]{\smicon{figures/icons/EmptyStar.pdf}}
\newcommand{\halfstar}[0]{\icon{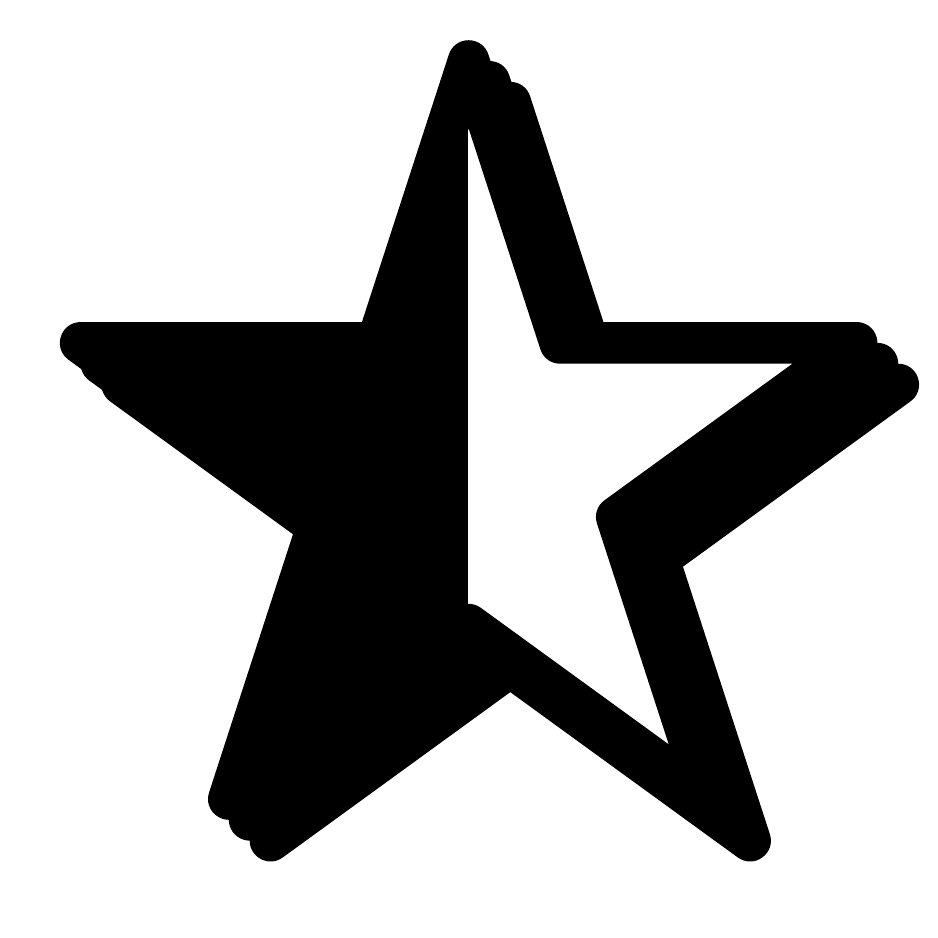}}
\newcommand{\fullstar}[0]{\icon{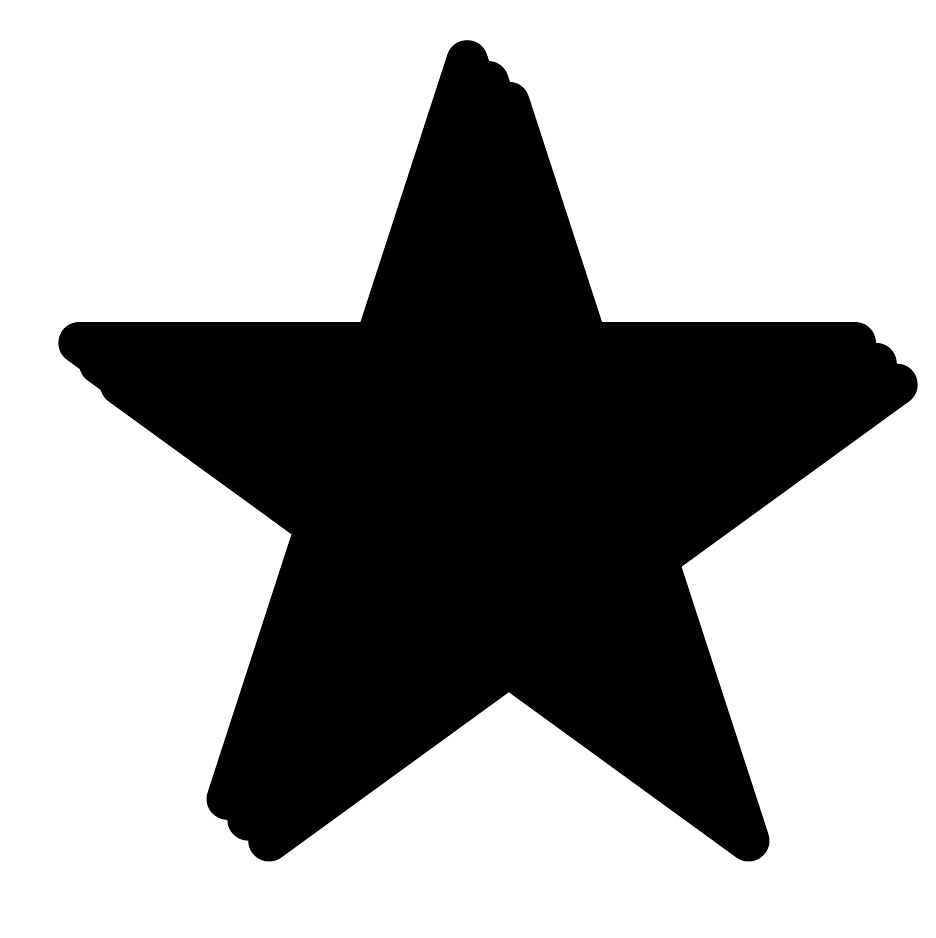}}
\newcommand{\smfullstar}[0]{\smicon{figures/icons/FullStar.pdf}}
\newcommand{\indoor}{\icon{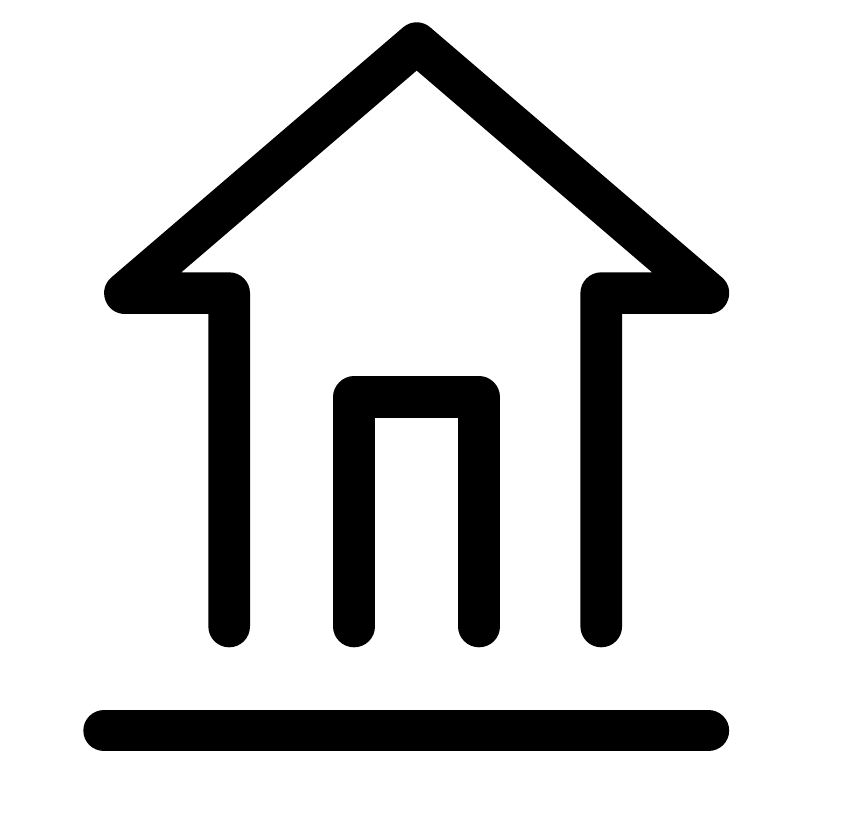}}
\newcommand{\outdoor}{\icon{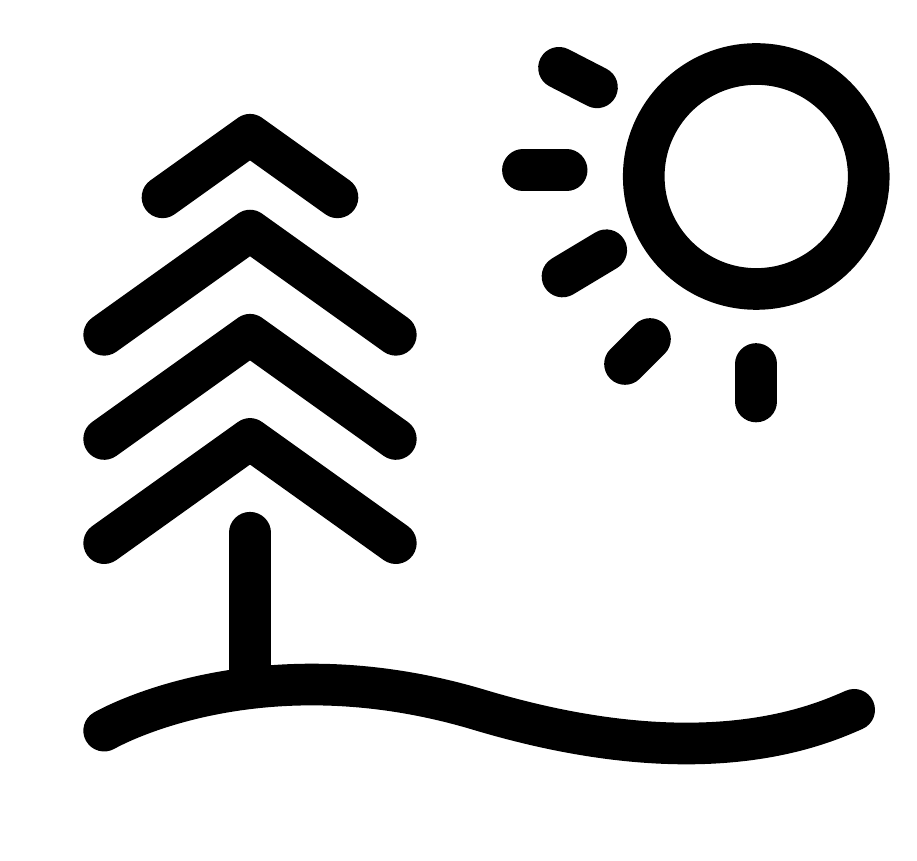}}
\newcommand{\indooroutdoor}{\indoor$+$\outdoor}
\newcommand{\robotnav}{\icon{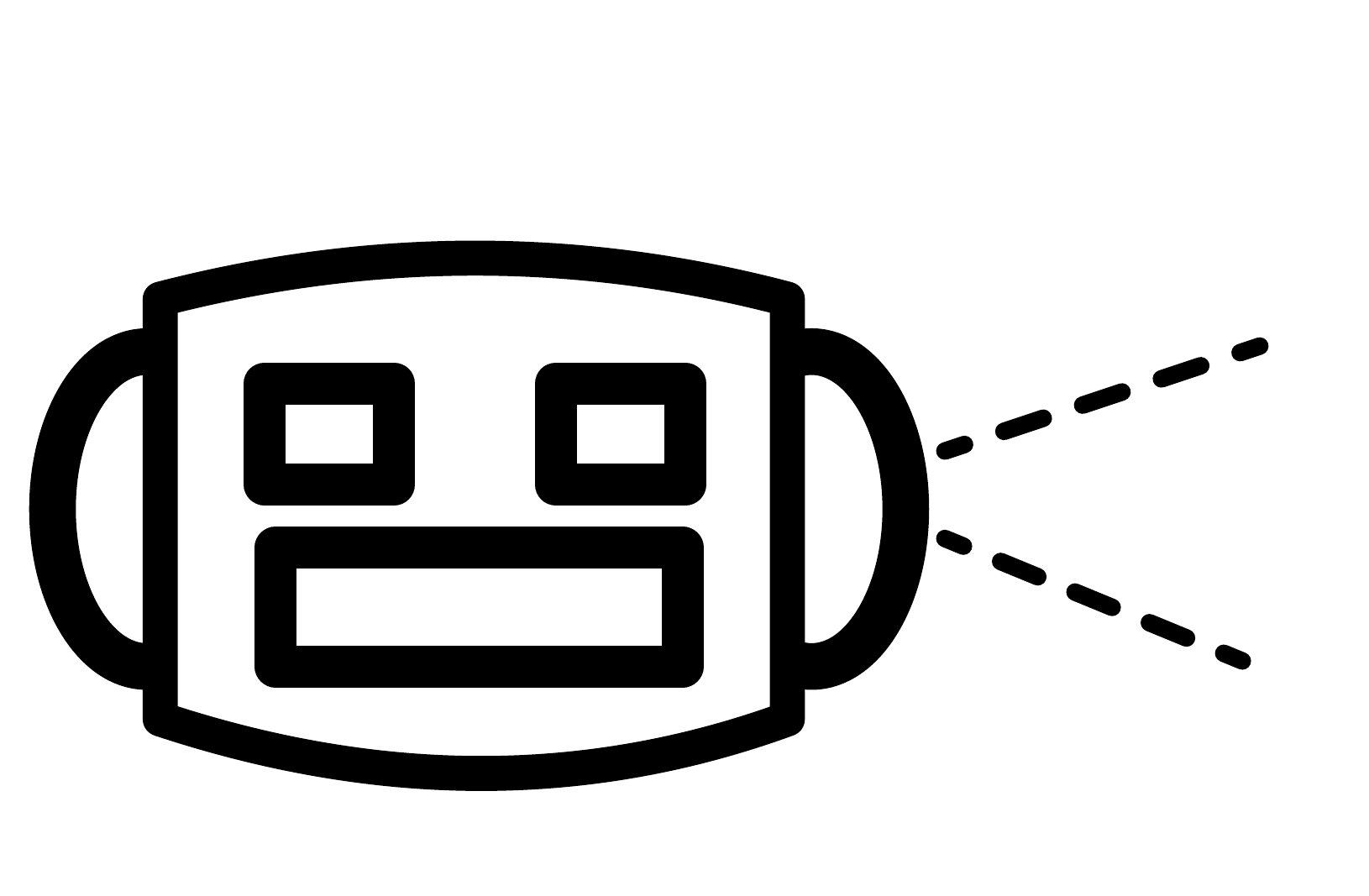}}
\newcommand{\selfdriving}{\icon{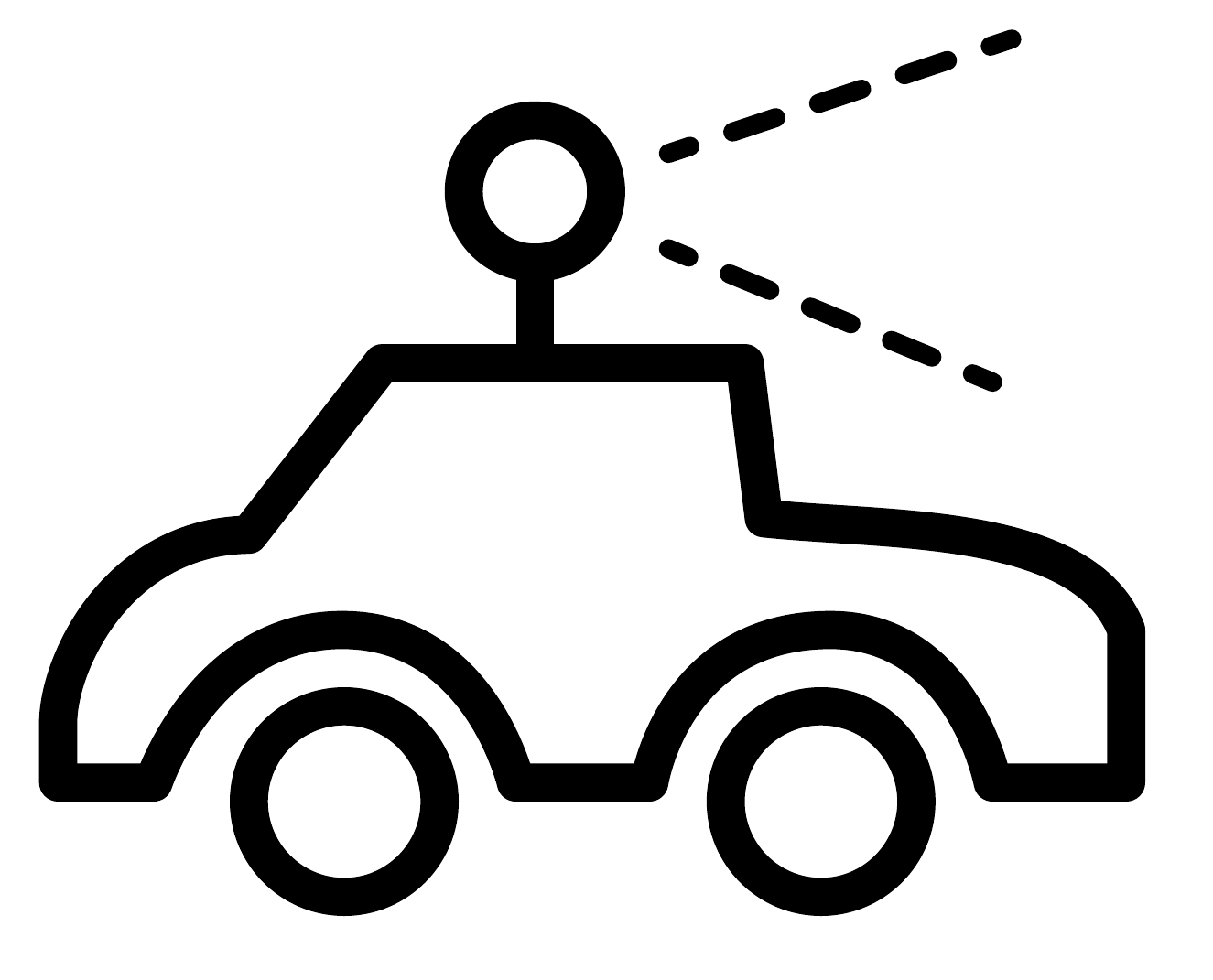}}
\newcommand{\videoseg}{V.O.S.}
\newcommand{\egonav}{\icon{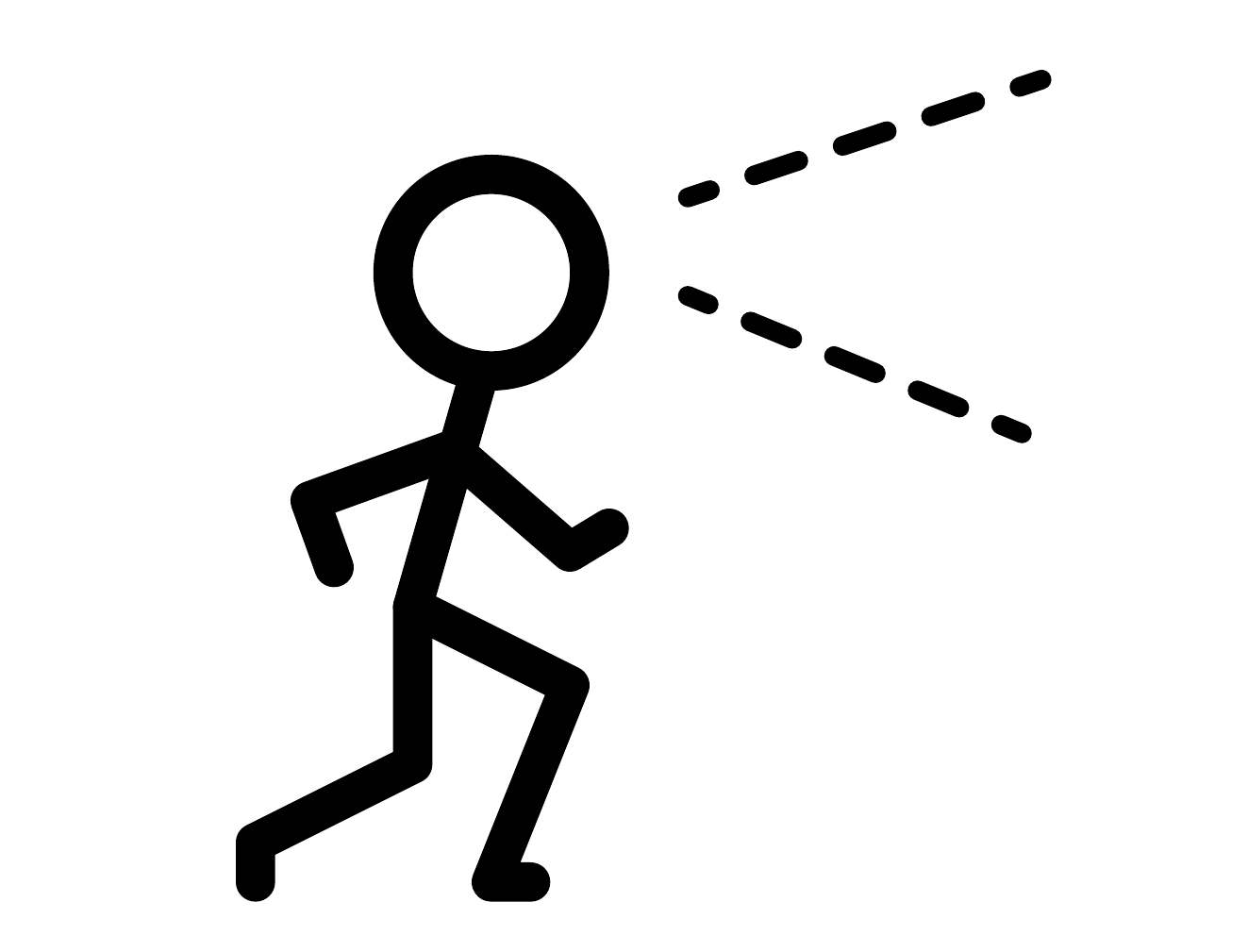}}
\newcommand{\stereo}{\icon{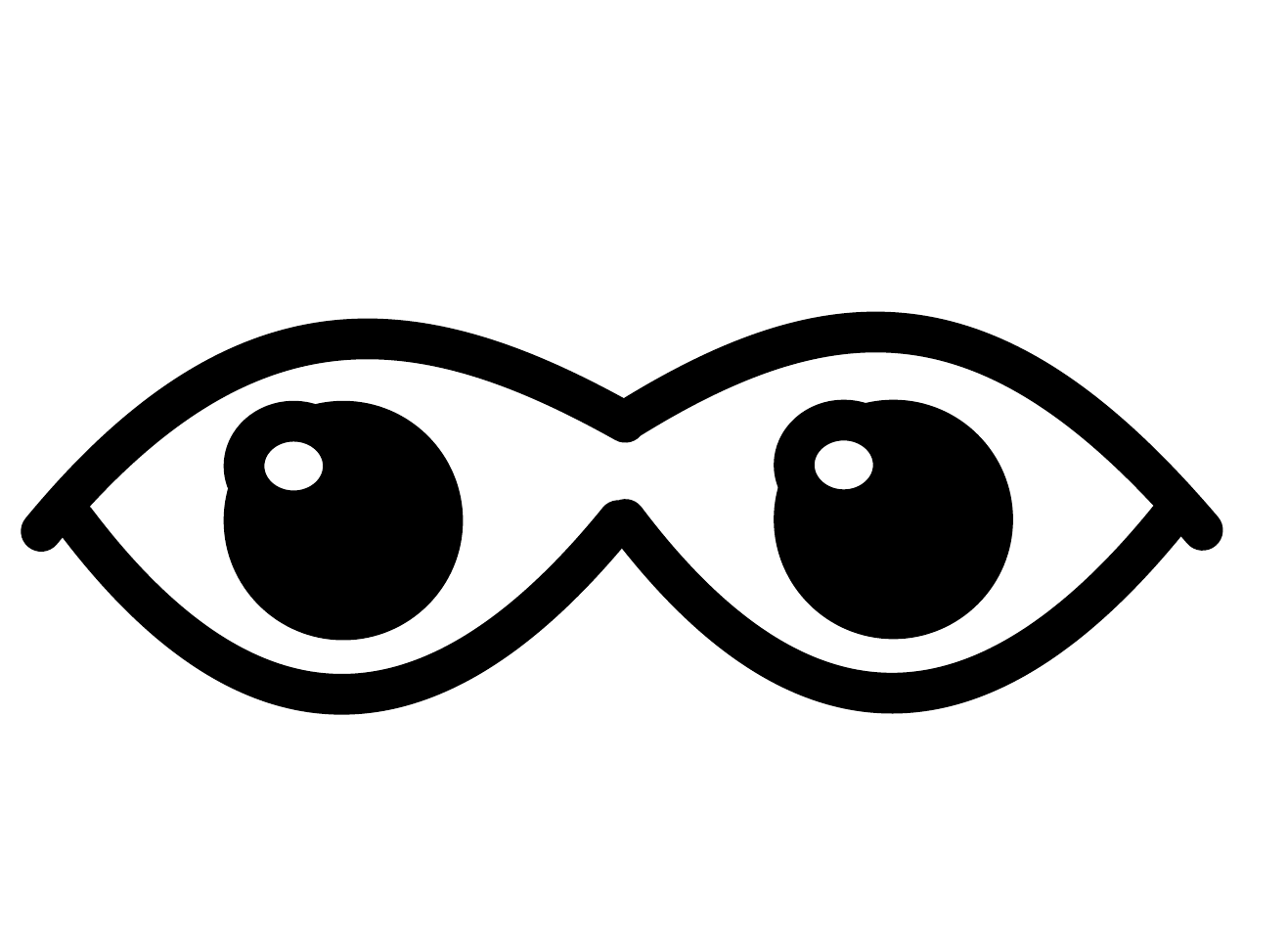}}
\newcommand{\pedestriandetection}{\icon{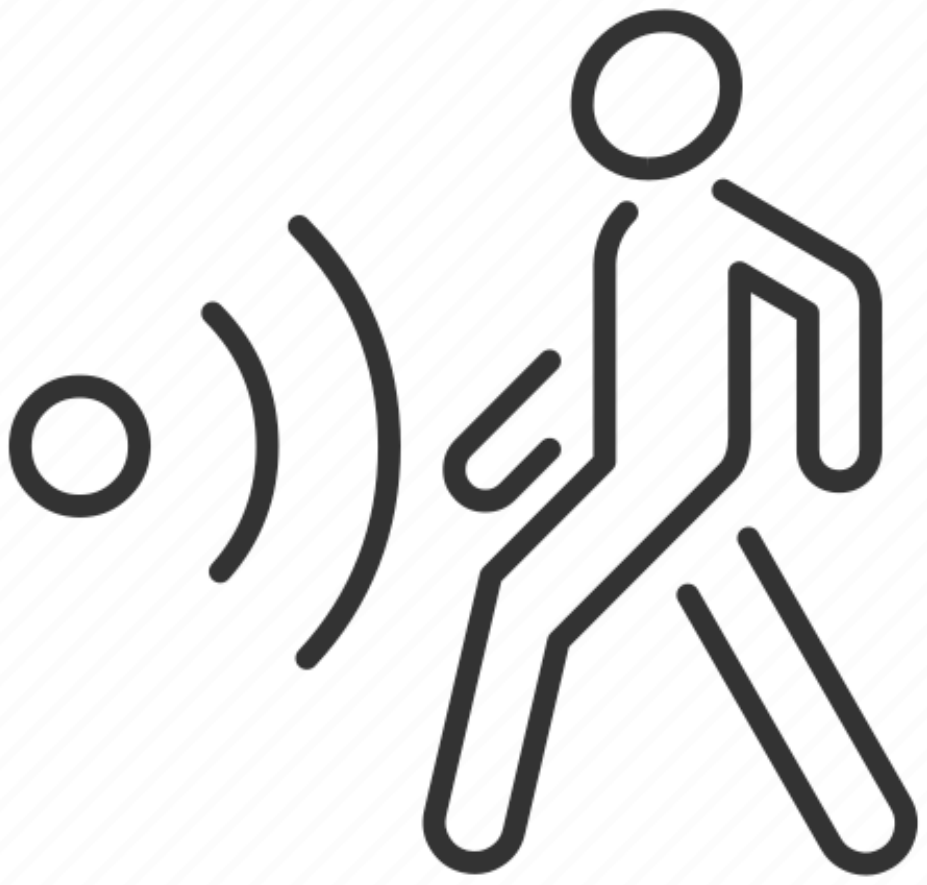}}

\centering

\begin{tabular}{ccccccc}
\toprule

\textbf{Dataset} & \textbf{Domain} & \textbf{Environment} & \textbf{\# Frames}
& \begin{tabular}{@{}c@{}} \textbf{\# Seg} \\ \textbf{Masks}\end{tabular}
& \begin{tabular}{@{}c@{}} \textbf{\# Depth} \\ \textbf{Maps}\end{tabular}\\
\midrule

SCAND \cite{karnan2022socially}
& \robotnav 
& \begin{tabular}{@{}c@{}} \indooroutdoor \\ \stereo \end{tabular}
& \begin{tabular}{@{}c@{}} $\sim$522 \\ minutes\end{tabular}
&
& \begin{tabular}{@{}c@{}} $\sim$522 \\ minutes\end{tabular}\\

MuSoHu \cite{nguyen2023toward}
& \robotnav 
& \begin{tabular}{@{}c@{}} \indooroutdoor \\ \stereo \end{tabular}
& \begin{tabular}{@{}c@{}} $\sim$600 \\ minutes\end{tabular}
&
& \begin{tabular}{@{}c@{}} $\sim$600 \\ minutes\end{tabular}\\

\midrule

\begin{tabular}{@{}c@{}}Playing for Benchmark \\(Synthetic) \cite{richter2017playing}\end{tabular} & \selfdriving 
& \outdoor 
& 250K 
& \begin{tabular}{@{}c@{}} 250K \\ (Dense)\end{tabular}
&\\

\begin{tabular}{@{}c@{}}Cityscapes-DVPS  \cite{vip_deeplab}\end{tabular} 
& \selfdriving & \outdoor $+$\stereo 
& 3K
& \begin{tabular}{@{}c@{}} 3K \\ (Dense)\end{tabular}
& \begin{tabular}{@{}c@{}} 3K \\ (Dense)\end{tabular}\\

\begin{tabular}{@{}c@{}} KITTI-360 \cite{Liao2022PAMI} \end{tabular} 
& \selfdriving 
& \outdoor 
& 320K
& \begin{tabular}{@{}c@{}} 2x78K \\ (Dense) \end{tabular}
& \begin{tabular}{@{}c@{}} 2x78K \\ (Dense) \end{tabular}\\

\begin{tabular}{@{}c@{}} Panoptic-nuScenes \cite{nuscenes2019} \end{tabular}{} 
& \selfdriving & \outdoor  
& 1.4M 
& \begin{tabular}{@{}c@{}} 40K \\ (Dense) \end{tabular}
&\\

\begin{tabular}{@{}c@{}}Waymo Open Dataset\\-Panoramic \cite{mei2022waymo}\end{tabular} 
& \selfdriving 
& \outdoor  
& 390K &  100K &\\

A$^*3$D \cite{astar-3d}
& \selfdriving 
& \outdoor  
& 39K
& \begin{tabular}{@{}c@{}} 39K \\ (3D BBox) \end{tabular}
& \\

\begin{tabular}{@{}c@{}} ApolloScape-SceneParsing\\ \cite{apolloscape_arXiv_2018} \end{tabular}
& \selfdriving 
& \outdoor  
& 140K 
& \begin{tabular}{@{}c@{}} 140K \\ (Dense) \end{tabular}
& \begin{tabular}{@{}c@{}} 140K \\ (Dense) \end{tabular}\\

DDAD \cite{packnet}
& \selfdriving 
& \outdoor  
& 21K
& 
& \begin{tabular}{@{}c@{}} 21K \\ (Dense) \end{tabular}\\

\begin{tabular}{@{}c@{}} DOLPHINS\\(Synthetic) \cite{Mao_2022_ACCV} \end{tabular}
& \selfdriving 
& \outdoor  
& 42K 
& \begin{tabular}{@{}c@{}} 42K \\ (3D BBox) \end{tabular}
&\\

\begin{tabular}{@{}c@{}}Argoverse2 \\ Sensor Data \cite{wilson2023argoverse}\end{tabular} 
& \selfdriving 
& \outdoor  $+$ \stereo 
& \begin{tabular}{@{}c@{}} 1000 \\ (Videos) \end{tabular}
& 3D BBox &\\

\begin{tabular}{@{}c@{}} CamVid \cite{BROSTOW200988} \end{tabular} & \selfdriving  & \outdoor & \begin{tabular}{@{}c@{}} 5 \\ (Videos) \end{tabular} & \begin{tabular}{@{}c@{}} 700 \\ (Dense) \end{tabular} & \\

\midrule

MS-COCO \cite{lin2014microsoft} & O.D.S & \indooroutdoor & 328K & 328K &\\

\begin{tabular}{@{}c@{}} Youtube-VOS \cite{xu2018youtube} \end{tabular}
& \videoseg 
& \indooroutdoor  
& $\sim$20K
& \begin{tabular}{@{}c@{}} $\sim$4K \\ (Sparse) \end{tabular}
&\\

\begin{tabular}{@{}c@{}} DAVIS-2017 \cite{Caelles_arXiv_2019}\end{tabular} & \videoseg  & \indooroutdoor & 10K & \begin{tabular}{@{}c@{}} 10K \\ (Sparse) \end{tabular} & \\



\midrule
SideGuide \cite{9340734}
& \egonav
& \outdoor
& \begin{tabular}{@{}c@{}} 2x180K, \\ 312K\end{tabular}
& \begin{tabular}{@{}c@{}} 100K \\ (Sparse)\end{tabular}
& \begin{tabular}{@{}c@{}} 180K \\ (Dense)\end{tabular}\\

\textbf{SANPO-Real} (ours)
& \egonav 
& \outdoor $+$ \stereo 
& 2x617K
& \begin{tabular}{@{}c@{}} 112K \\ (Dense) \end{tabular}
& \begin{tabular}{@{}c@{}} 617K \\ (Dense) \end{tabular}\\

\textbf{SANPO-Synthetic} (ours)
& \egonav 
& \outdoor  
& 113K
& \begin{tabular}{@{}c@{}} 113K \\ (Dense) \end{tabular}
& \begin{tabular}{@{}c@{}} 113K \\ (Dense) \end{tabular}\\

\midrule

\multicolumn{6}{c}{}\\
\multicolumn{6}{c}{{\begin{tabular}{@{}c@{}} 
\robotnav$\quad$: Robot Navigation, \selfdriving$\quad$: Self-Driving, \egonav$\;\;$: Egocentric Navigation,\\
O.D.S: Object Detection \& Segmentation,
V.O.S: Video Object Segmentation\\
\indoor: Indoor, \outdoor: Outdoor, \stereo: Stereo \end{tabular}}}\\

\bottomrule

\end{tabular}
\caption{\label{table:datasets-comparison}\textbf{Dataset Comparison}. SANPO is a unique video dataset designed to address a gap in current offerings. Unlike existing datasets focused on self-driving vehicles or general video object segmentation (VOS), SANPO targets the specific challenges of egocentric human navigation. SANPO is a large-scale, challenging, and diverse dataset. It offers both real and synthetic data, with multi-view stereo data included in the real component.}
\end{table*}

While many outdoor video datasets exist for tasks like robot navigation, autonomous driving, and video segmentation (see Table ~\ref{table:datasets-comparison}), SANPO fills a crucial gap.  To our knowledge, it is the only dataset providing both real and synthetic data with panoptic labels and depth maps specifically designed for human-centric egocentric navigation research.

Figure~\ref{fig:synth-vs-real} provides a visual comparison between SANPO-Real and SANPO-Synthetic.
\begin{figure}[h]%
    
    \centering%

    \includegraphics[bb=0 0 18.8cm 5.28cm, width=0.8\linewidth]{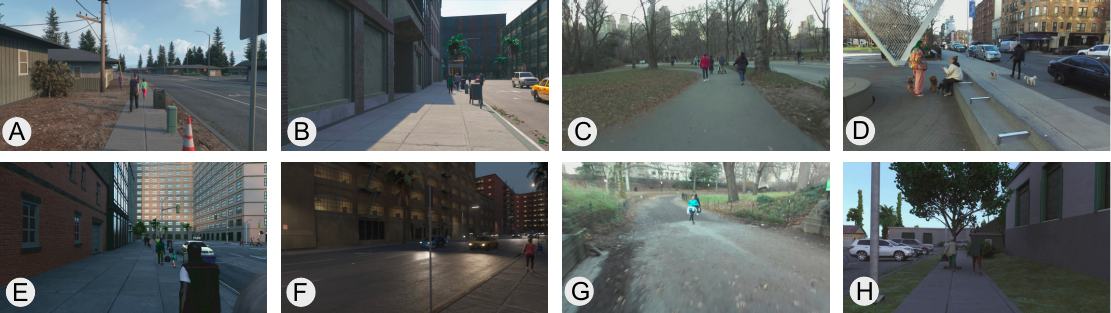}%
    \caption{\textbf{SANPO Synthetic vs real.} A sample of SANPO-Real and SANPO-Synthetic data. \emph{How quickly can you tell which of these images is synthetic?} Answer key in base64: `\texttt{c3ludGg6IEFCRUZILCByZWFsOiBDREc=}'}%
    \label{fig:synth-vs-real}%
\end{figure}

\subsubsection{Additional Statistics}
\label{ref:additional-statistics}
\paragraph{Pedestrian density}
\begin{figure}[th]
\includegraphics[width=\linewidth]{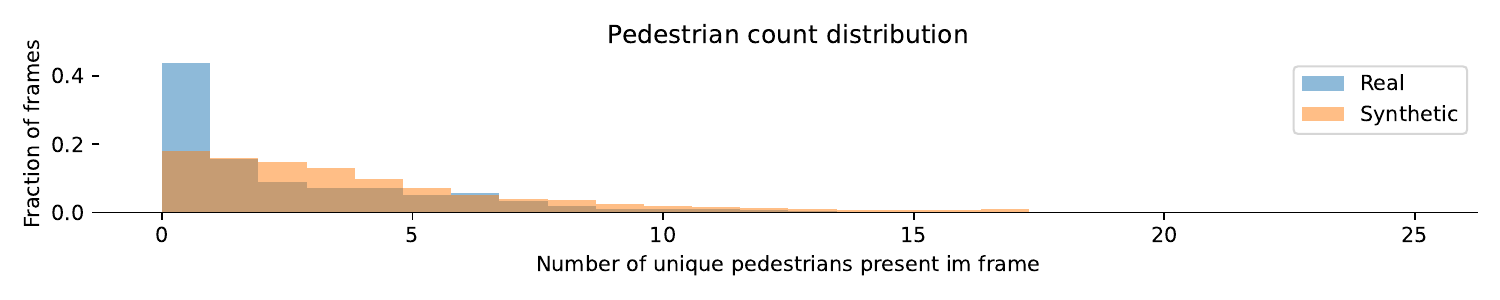}
\caption{\label{fig:pedestrians}\textbf{Distribution of pedestrians} in SANPO-Real and SANPO-Synthetic. SANPO-Real frequently features images with no pedestrians, but pedestrians appear in almost all frames of SANPO-Synthetic, and in greater quantities.}
\end{figure}
\paragraph{Additional Attributes of SANPO-Synthetic's Segmentation Annotations}
\begin{itemize}
    \item Instance Density: Over half of the frames have $\geq60$ unique instances, with a sixth having $\geq150$.
    \item Small Objects: 80\% of object masks have less than $32^2$ pixels, significantly more than SANPO-Real (8.1\%).
\end{itemize}

\subsection{Data Annotation}
\label{appendix:data_annotatio}
\subsubsection{Session Attributes}
\label{appendix:session_attributes}
Each real session is annotated with the following high level attributes.
\begin{enumerate}
    \item Human Traffic
    \begin{enumerate}
        \item Low
        \item Moderate
        \item Heavy
    \end{enumerate}
 
    \item Vehicular Traffic
    \begin{enumerate}
        \item Low
        \item Moderate
        \item Heavy
    \end{enumerate}
 
    \item Animal Traffic
    \begin{enumerate}
        \item Low
        \item Moderate
        \item Heavy
    \end{enumerate}
 
    \item Number of Obstacles
    \begin{enumerate}
        \item Low
        \item Moderate
        \item Heavy
    \end{enumerate}

    \item Environment Type
    \begin{enumerate}
      \item Urban
      \item Suburban
      \item Rural
      \item Park
      \item Road Junction
      \item Open Terrain
      \item Open Space
      \item Indoor
    \end{enumerate}

    \item Weather Condition
    \begin{enumerate}
      \item Sunny
      \item Cloudy
      \item Rainy
      \item Snowy
    \end{enumerate}
    
    \item Visibility
    \begin{enumerate}
      \item High
      \item Medium
      \item Low
    \end{enumerate}
    
    \item Motion Type 
    \begin{enumerate}
      \item Walking
      \item Jogging
      \item Running
    \end{enumerate}
    
    \item Elevation Change
    \begin{enumerate}
      \item Flat
      \item Uphill
      \item Downhill
      \item Stairs
    \end{enumerate}
    
    \item Ground Appearances
    \begin{enumerate}
      \item Light Gray
      \item Dark Gray
      \item Pavers
      \item Color
      \item Terrain
      \item Gravel
      \item Sand
    \end{enumerate}
    
    \item Motion Blur
    \begin{enumerate}
      \item Low
      \item Medium
      \item High
    \end{enumerate}
    
    \item Rare Events
\end{enumerate}

\subsubsection{SANPO Taxonomy}
\label{appendix:sanpo-taxonomy}
SANPO taxonomy labels with \textit{stuff} or \textit{thing} distinction.

\begin{enumerate}\addtocounter{enumi}{-1}
    \item  unlabeled \({\rightarrow}\) stuff
    \item  road \({\rightarrow}\) stuff
    \item  curb \({\rightarrow}\) stuff
    \item  sidewalk \({\rightarrow}\) stuff
    \item  guard rail/road barrier \({\rightarrow}\) stuff
    \item  crosswalk \({\rightarrow}\) thing
    \item  paved trail \({\rightarrow}\) stuff
    \item  building \({\rightarrow}\) stuff
    \item  wall/fence \({\rightarrow}\) stuff
    \item  hand rail \({\rightarrow}\) stuff
    \item  opening-door \({\rightarrow}\) thing
    \item  opening-gate \({\rightarrow}\) thing
    \item  pedestrian \({\rightarrow}\) thing
    \item  rider \({\rightarrow}\) thing
    \item  animal \({\rightarrow}\) thing
    \item  stairs \({\rightarrow}\) thing
    \item  water body \({\rightarrow}\) stuff
    \item  other walkable surface \({\rightarrow}\) stuff
    \item  inaccessible surface \({\rightarrow}\) stuff
    \item  railway track \({\rightarrow}\) stuff
    \item  obstacle \({\rightarrow}\) thing
    \item  vehicle \({\rightarrow}\) thing
    \item  traffic sign \({\rightarrow}\) thing
    \item  traffic light \({\rightarrow}\) thing
    \item  pole \({\rightarrow}\) thing
    \item  bus stop \({\rightarrow}\) thing
    \item  bike rack \({\rightarrow}\) thing
    \item  sky \({\rightarrow}\) stuff
    \item  tree \({\rightarrow}\) thing
    \item  vegetation \({\rightarrow}\) stuff
    \item  terrain \({\rightarrow}\) stuff
\end{enumerate}

\subsubsection{Segmentation Annotation Process}
\label{appendix:seg_process}
In this section we describe the segmentation annotation process for SANPO-Real. We divide each video into 30-second sub-videos (note: most videos are only 30 seconds long, resulting in a single sub-video),
then we annotate every sixth frame (0-6-12-...), for a total of 90 frames per sub-video.
To enhance efficiency and accuracy of human annotation, we employ two key techniques:
\begin{itemize}
    \item Cascaded Annotation: To manage our extensive taxonomy, we divide all labels into five mutually exclusive subsets containing commonly co-occurring labels. Each sub-video is annotated in a temporally consistent manner across these subsets in a carefully determined optimal order\footnote{We experimented with various combinations to refine this approach.}. When annotating a subset, previously annotated regions are frozen and displayed to the annotator, thus increasing their speed and improving boundary precision. The final subset includes all labels, ensuring that any regions missed in previous subsets are annotated.
    
    \item AOT based Propagation:  We leverage AOT \cite{yang2021aot} to propagate masks from human-annotated frames to the intermediate unannotated frames. We track whether each frame is human-annotated or machine-propagated, and this information is included alongside the provided annotations. Figure~\ref{fig:consistent_segmentation_annotation} visually demonstrates this process, showing human-annotated frames and their machine-propagated counterparts.
    
\end{itemize}
\begin{figure}
    
    \centering
    \subfloat{{\includegraphics[width=0.4\linewidth]{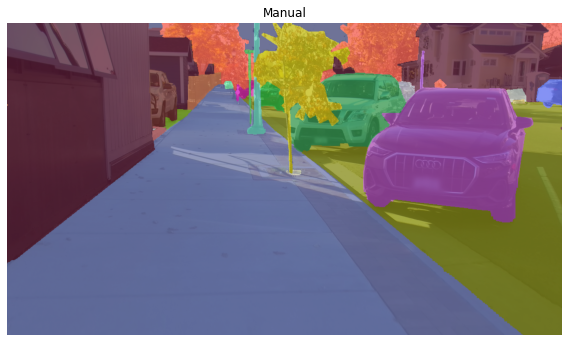}}}%
    \subfloat{{\includegraphics[width=0.4\linewidth]{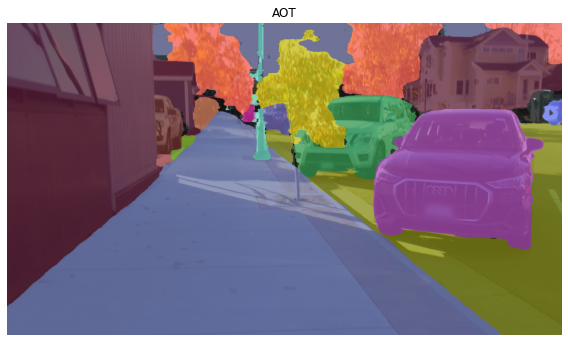}}}%

    \subfloat{{\includegraphics[width=0.4\linewidth]{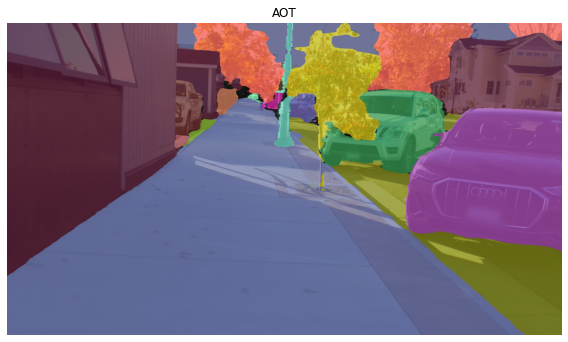}}}%
    \subfloat{{\includegraphics[width=0.4\linewidth]{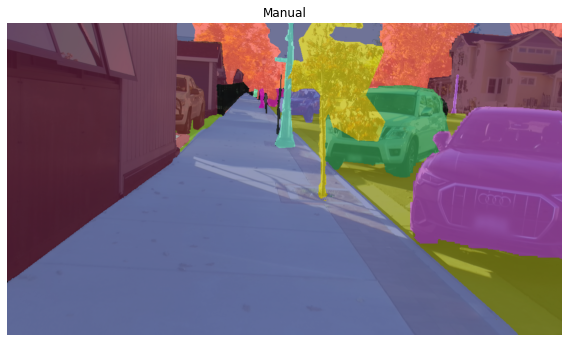}}}%
    

    \caption{
\textbf{Temporally Consistent Segmentation Annotation.}\label{fig:consistent_segmentation_annotation}
Our annotation process ensures temporal consistency across both human-annotated and machine-propagated masks. Compare the first and last columns of the figure to see this consistency. Most propagated masks are accurate, with occasional failures for thin objects like trees (yellow) and poles (cyan).}

\end{figure}

This process resulted in 18,787 human-annotated frames and 93,981 machine-propagated frames.

\paragraph{Evaluating AOT-Based Propagation Accuracy}
\label{appendix:aot-accuracy}
To evaluate the accuracy of AOT-based propagation for segmentation annotations, we performed the following analysis. We considered human-annotated frames (0-6-12-...), propagated segmentation masks to every other frame (6-18-...) using AOT, and compared these propagated masks to the corresponding human-annotated ground truth (GT) to calculate a propagation score. Since the motion gap between these frames is significant, this method provides a conservative estimate (lower bound) of the propagation error.
In accordance with the video object segmentation (VOS) literature \cite{perazzi2016benchmark, yang2021aot, ding2023mose}, we used region similarity \textit{J} and contour accuracy \textit{F} as evaluation metrics. The mean \textit{J\&F} score for SANPO-Real is 0.892, demonstrating a strong lower bound on the accuracy of machine-propagated masks.

\paragraph{Detailed and Accurate Segmentation Annotation}
Our dataset captures rich details, including high-quality semantic masks for even the smallest objects (see Fig.~\ref{fig:small_instances_mask} for examples).
\begin{figure}
\centering%
\includegraphics[height=0.8\linewidth, width=0.8\linewidth]{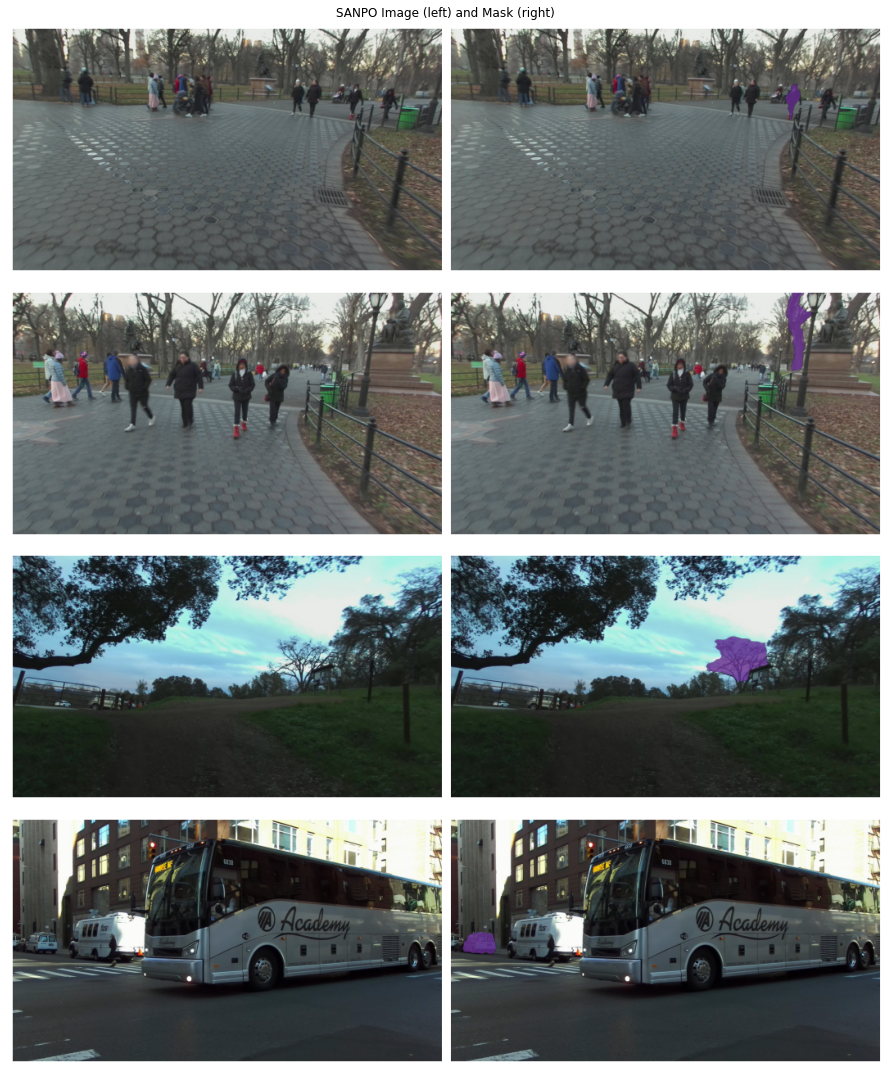}%
\caption{SANPO's detailed annotation include masks for even the smallest objects (highlighted in purple, right column).}
\label{fig:small_instances_mask}
\end{figure}

\subsection{Benchmarks}
\subsubsection{Zero Shot Evaluation}
\label{appendix:cityscapes-to-sanpo}
\textbf{Cityscapes-19 -> SANPO Mapping}\\
To ensure a fair comparison, we map Cityscapes-19 labels to SANPO labels wherever possible. Below mapping from Cityscapes-19 to SANPO taxonomy:
\begin{multicols}{2}
\begin{enumerate}
\item road \({\rightarrow}\)  road
\item sidewalk \({\rightarrow}\)  sidewalk
\item building \({\rightarrow}\)  building
\item wall \({\rightarrow}\)  wall/fence
\item fence \({\rightarrow}\)  wall/fence
\item pole \({\rightarrow}\)  pole
\item traffic light \({\rightarrow}\)  traffic light
\item traffic sign \({\rightarrow}\)  traffic sign
\item vegetation \({\rightarrow}\)  vegetation
\item terrain \({\rightarrow}\)  terrain
\item sky \({\rightarrow}\)  sky
\item person \({\rightarrow}\)  pedestrian
\item rider \({\rightarrow}\)  rider
\item car \({\rightarrow}\)  vehicle
\item truck \({\rightarrow}\)  vehicle
\item bus \({\rightarrow}\)  vehicle
\item train \({\rightarrow}\)  vehicle
\item motorcycle \({\rightarrow}\)  vehicle
\item bicycle \({\rightarrow}\)  vehicle
\end{enumerate}
\end{multicols}

For all SANPO labels without an appropriate mapping from Cityscapes-19, we treat the corresponding pixels as unlabeled and exclude them from the mIoU metric computation in the zero-shot semantic segmentation evaluation. The following SANPO labels were excluded:
\begin{multicols}{2}
\begin{enumerate}
\item curb
\item guard rail/road barrier
\item crosswalk
\item paved trail
\item hand rail
\item opening-door
\item opening-gate
\item animal
\item stairs
\item water body
\item other walkable surface
\item inaccessible surface
\item railway track
\item obstacle
\item bus stop
\item bike rack
\item tree
\end{enumerate}
\end{multicols}

\textbf{Zero-shot Mask2Former Evaluation}\\
We also evaluated the Mask2Former Swin-L model \cite{cheng2021mask2former} in the zero-shot setting. Despite its strong performance on Cityscapes (mIoU 0.833), it achieved lower scores on SANPO-Real (0.417) and SANPO-Synthetic (0.476). 
Fig.~\ref{fig:mask2former-eval} offers a qualitative assessment on SANPO samples and Table~\ref{table:mask2former-per-class} provides a class-wise mIoU breakdown.
\begin{figure}[h]
    \centering
    \subfloat{{\includegraphics[bb=0 0 80cm 46cm, width=.4\linewidth]{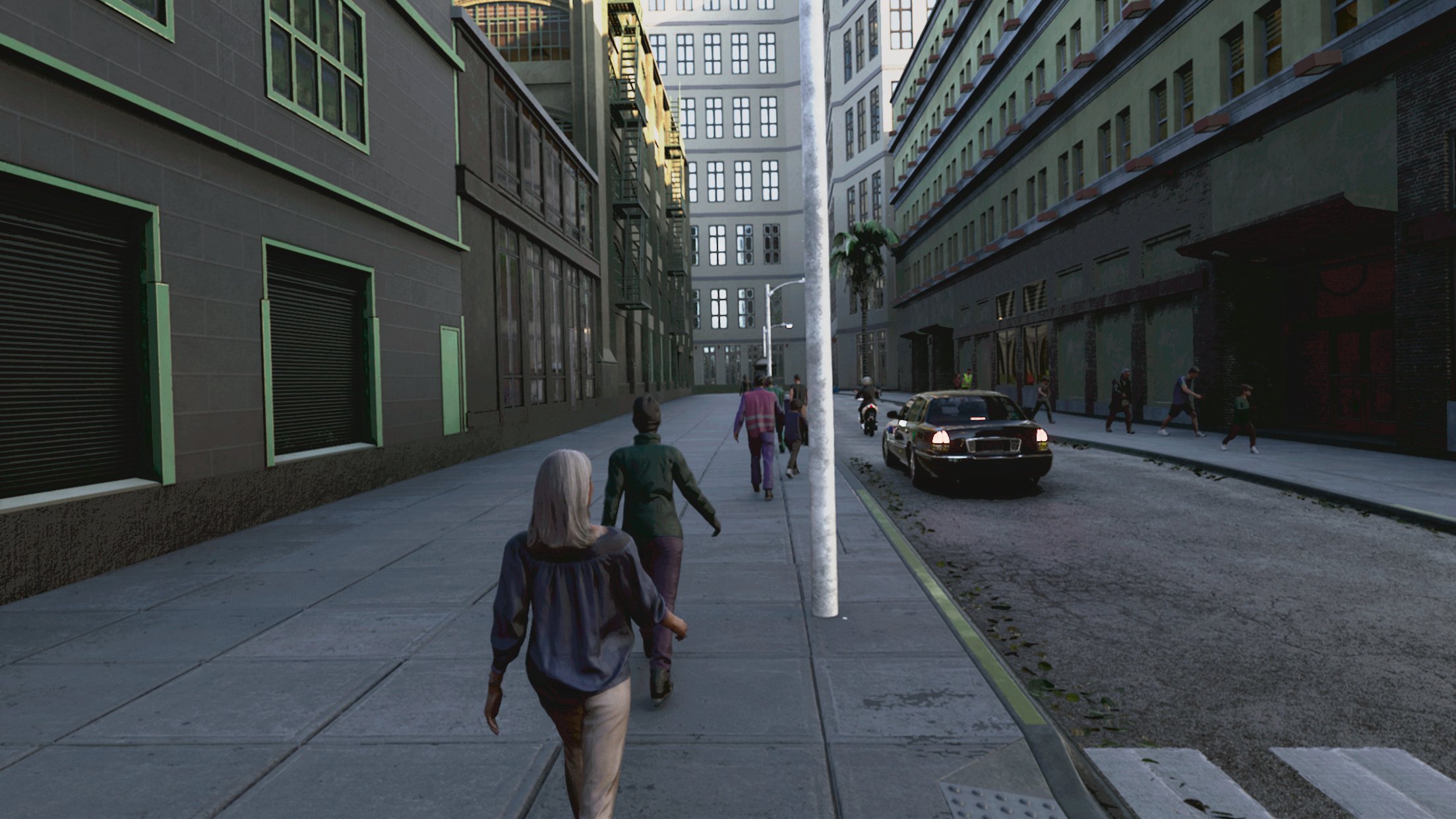}}}%
    \subfloat{{\includegraphics[bb=0 0 58cm 30cm, width=.4\linewidth]{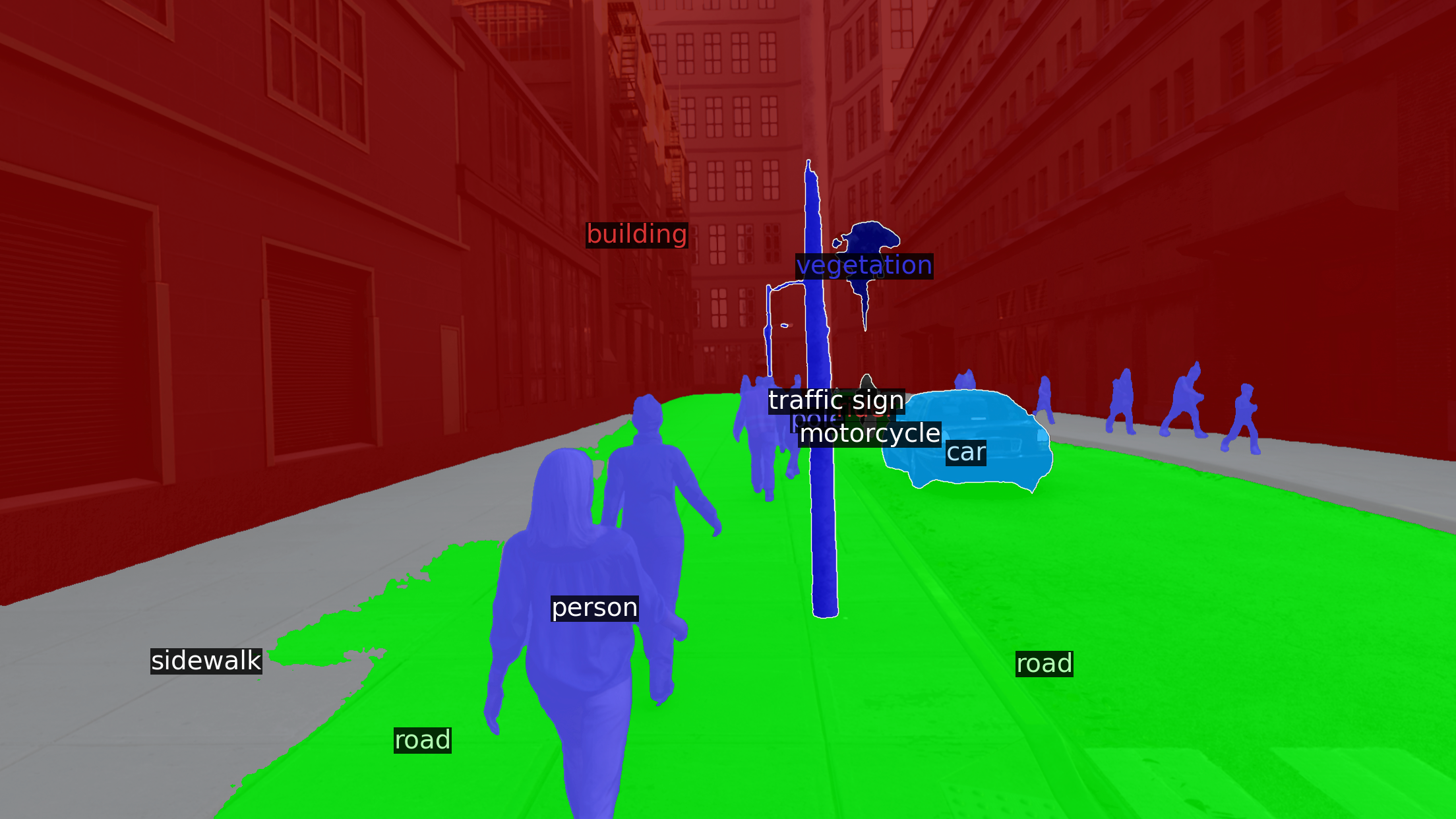}}}%
    
    \subfloat{{\includegraphics[bb=0 0 80cm 46cm, width=.4\linewidth]{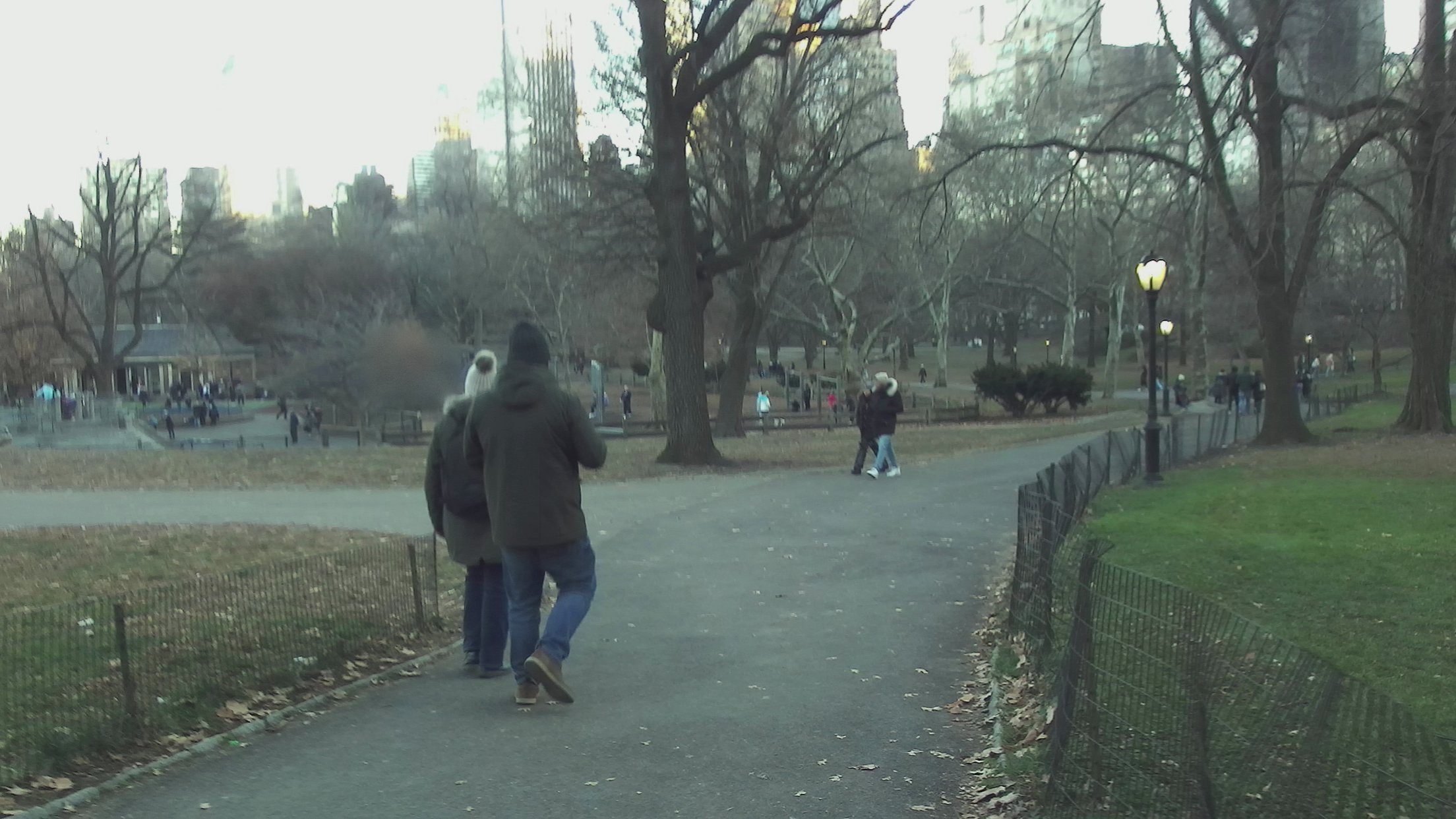}}}%
    \subfloat{{\includegraphics[bb=0 0 58cm 30cm, width=.4\linewidth]{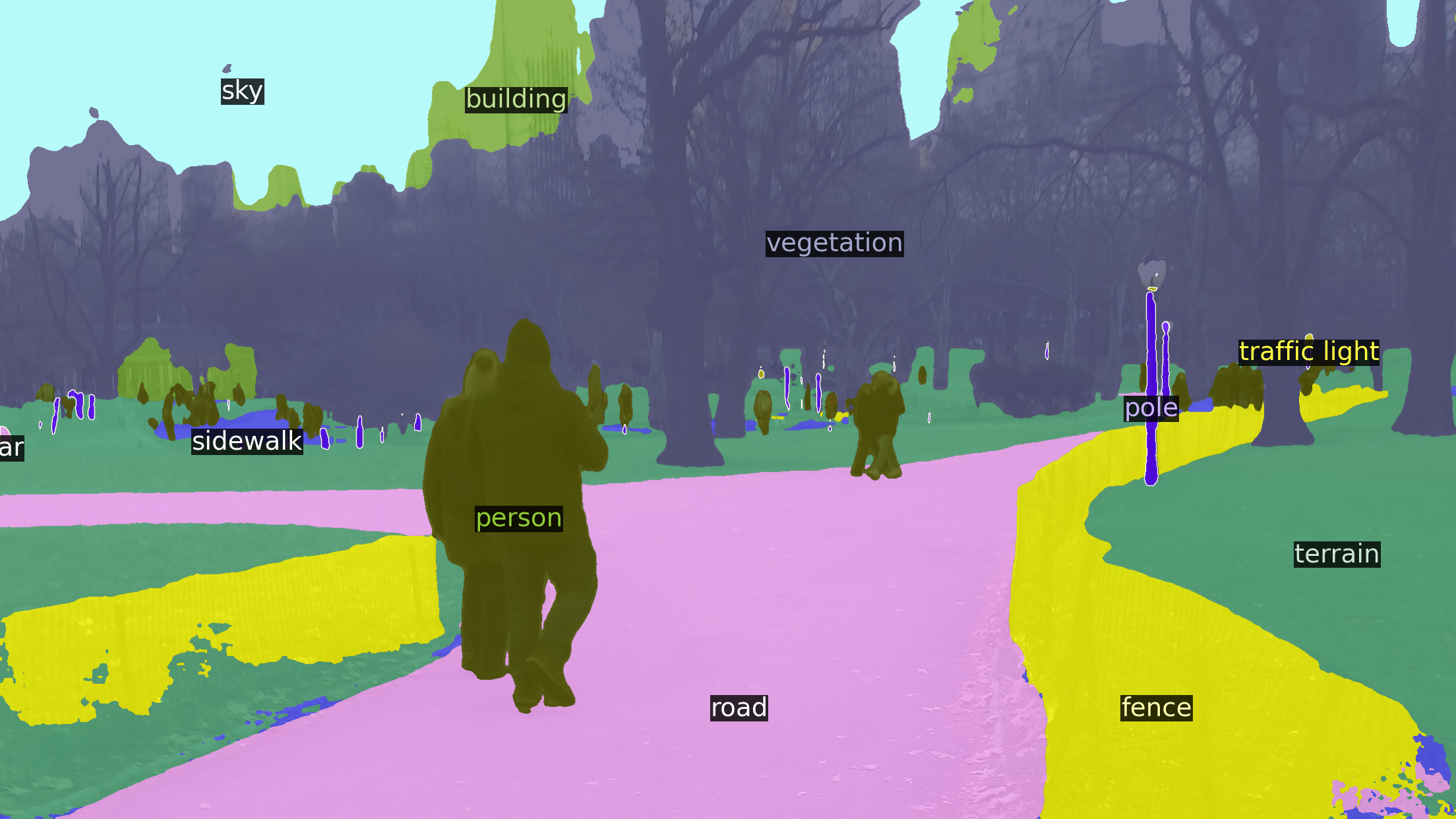}}}%
    \caption{\textbf{Highlight on Domain Gap.} 
    Egocentric navigation models must accurately differentiate between road (a not safe to walk surface) and sidewalk (a safe to walk surface). Mask2Former trained on Cityscapes dataset, similar to the Kmax-Deeplab models, struggles with this distinction on SANPO samples (top: synthetic, bottom: real). This, along with Table~\ref{table:zero-shot}, underscores the limited transferability of such datasets to human-centric navigation tasks.
    This visualization is generated using the Mask2Former tool \cite{cheng2021mask2former}.} 
    \label{fig:mask2former-eval}%
    
\end{figure}

\subsubsection{SANPO Benchmark}
\label{appendix:sanpo-benchmark-setup}
To ensure fairness and reproducibility, we maintained the following training setup:
\begin{enumerate}
    \item \textbf{Encoder Pretraining:} All encoders pretrained on ImageNet \cite{russakovsky2015imagenet}.
    \item \textbf{Datasets Used:} Only SANPO train sets.
    \item \textbf{Resizing:} Data resized to 1089x1921 (height x width), padding used to maintain aspect ratio.
    \item \textbf{Hyperparameters:} Standard values as defined in \cite{deeplab2_2021}.
    \item \textbf{Training Budget:} 60,000 steps with a batch size of 32 (doubled for Synthetic-to-Real domain adaptation fine-tuning experiments (\text{->} and \text{+} rows in Table~\ref{table:train-synthetic-to-real})). Approximate epochs for reference:
        \begin{itemize}
            \item Cityscapes Panoptic Segmentation: ~645
            \item SANPO-Real Panoptic Segmentation: ~21
            \item SANPO-Real (Human GT Only) Panoptic Segmentation: ~129
            \item SANPO-Synthetic Panoptic Segmentation and Depth Estimation: ~21
            \item SANPO-Real Depth Estimation: ~4
        \end{itemize}
\end{enumerate}

\begin{table}[h]

\centering
\scalebox{0.87}{
\begin{tabular}{ccc} 
\toprule

& \multicolumn{2}{c}{mIoU}
\\\cmidrule(lr){1-3}

Mapped SANPO Label & SANPO-Real  & SANPO-Synthetic
\\\cmidrule(lr){1-3}

road          & 0.255 & 0.407  \\ 
sidewalk      & 0.120 & 0.262 \\ 
building      & 0.642 & 0.934 \\ 
wall/fence    & 0.448 & 0.087 \\
pedestrian    & 0.679 & 0.878 \\
rider         & 0.271 & 0.247    \\
vehicle       & 0.658 & 0.817    \\
traffic sign  & 0.212 & 0.240    \\
traffic light & 0.127 & 0.344    \\
pole          & 0.310 & 0.586    \\
sky           & 0.658 & 0.919    \\
vegetation    & 0.654 & 0.303    \\
terrain       & 0.394 & 0.166
\\\cmidrule(lr){1-3}

\textbf{Average} & \textbf{0.417} & \textbf{0.476} \\

\bottomrule

\end{tabular}
}
\caption{\textbf{Mask2Former Zero-Shot Evaluation:} Per label breakdown of mIoU on the Mask2Former (Cityscapes) zero-shot experiment.}
\label{table:mask2former-per-class}
\end{table}


\begin{figure*}[t]
\includegraphics[width=1\linewidth]{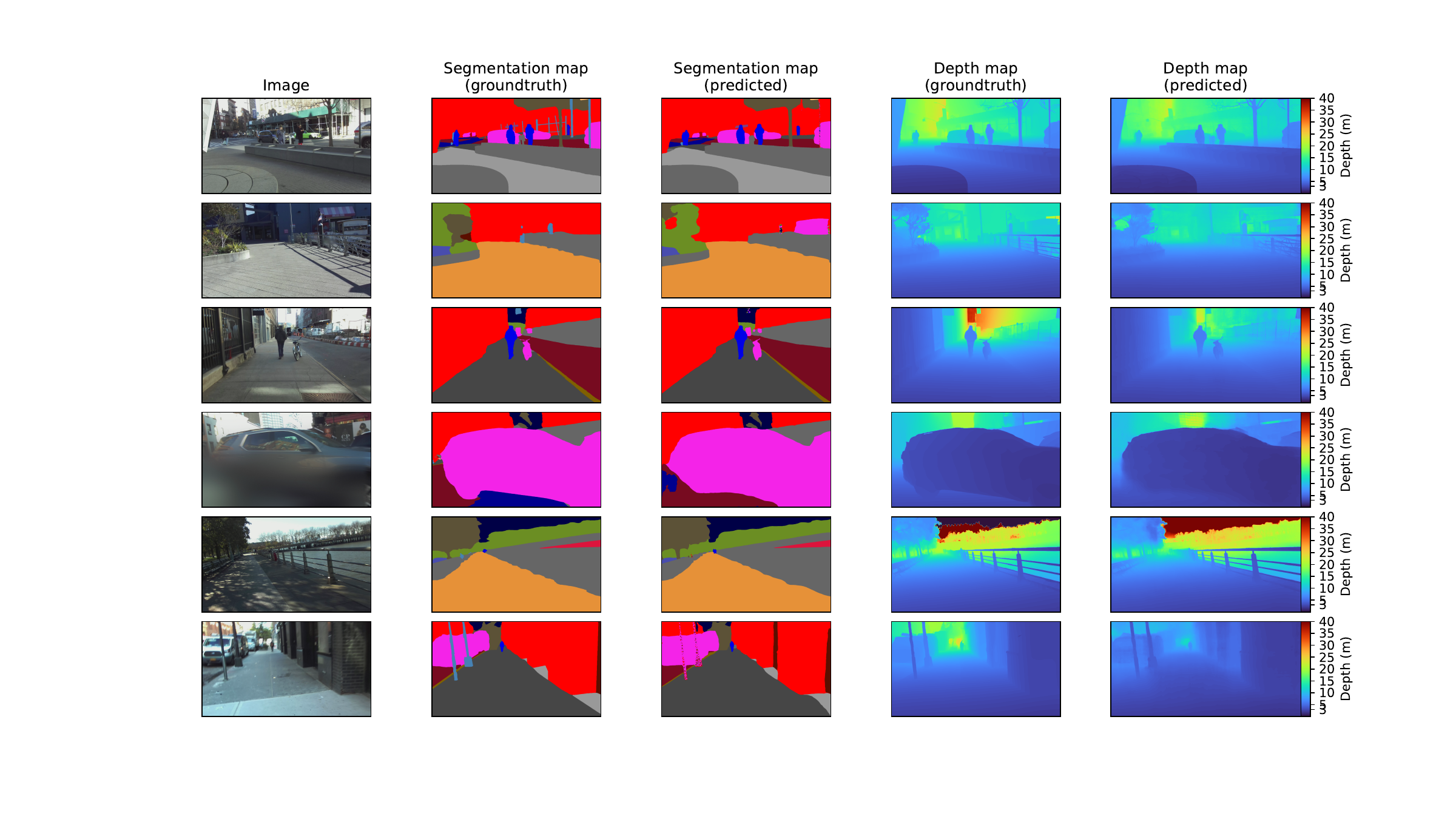}
\caption{\textbf{Qualitative examples on SANPO.} Showing left to right: image, groundtruth \& predicted segmentation maps, and groundtruth \& predicted metric depth maps.}
\centering
\label{fig:qual}
\end{figure*}
\subsection{SANPO Dense Prediction Qualitative Examples}
We show some example images in Fig.~\ref{fig:qual}, as well as ground truth and predicted segmentation maps from kMax-Deeplab and ground truth and predicted depth maps from Binsformer.

\subsection{Application}
\subsubsection{SANPO -> Accessibility Mapping}
\label{appendix:accessibility-mapping}
\emph{"safe to walk"} (e.g. sidewalk) and \emph{"not safe to walk"} (e.g. road, which is for vehicles) are ground surfaces.
\begin{enumerate}
\item unlabeled \({\rightarrow}\) not safe to walk
\item road \({\rightarrow}\) not safe to walk
\item curb \({\rightarrow}\) not safe to walk
\item sidewalk \({\rightarrow}\) safe to walk
\item guard rail/road barrier \({\rightarrow}\) obstacle
\item crosswalk \({\rightarrow}\) safe to walk
\item paved trail \({\rightarrow}\) safe to walk
\item building \({\rightarrow}\) obstacles
\item wall/fence \({\rightarrow}\) obstacles
\item hand rail \({\rightarrow}\) obstacles
\item opening-door \({\rightarrow}\) obstacles
\item opening-gate \({\rightarrow}\) obstacles
\item pedestrian \({\rightarrow}\) obstacles
\item rider \({\rightarrow}\) obstacles
\item animal \({\rightarrow}\) obstacles
\item stairs \({\rightarrow}\) safe to walk
\item water body \({\rightarrow}\) not safe to walk
\item other walkable surface \({\rightarrow}\) safe to walk
\item inaccessible surface \({\rightarrow}\) not safe to walk
\item railway track \({\rightarrow}\) not safe to walk
\item obstacle \({\rightarrow}\) obstacles
\item vehicle \({\rightarrow}\) obstacles
\item traffic sign \({\rightarrow}\) obstacles
\item traffic light \({\rightarrow}\) obstacles
\item pole \({\rightarrow}\) obstacles
\item bus stop \({\rightarrow}\) obstacles
\item bike rack \({\rightarrow}\) obstacles
\item sky \({\rightarrow}\) not safe to walk
\item tree \({\rightarrow}\) obstacles
\item vegetation \({\rightarrow}\) obstacles
\item terrain \({\rightarrow}\) safe to walk
\end{enumerate}

    {\small
    \bibliographystyle{ieee_fullname}
    \bibliography{main}
    }

  \else

    \maketitle
    \begin{abstract}





Vision is essential for human navigation. The World Health Organization (WHO) estimates that 43.3 million people were blind in 2020, and this number is projected to reach 61 million by 2050. Modern scene understanding models could empower these people by assisting them with navigation, obstacle avoidance and visual recognition capabilities. The research community needs high quality datasets for both training and evaluation to build these systems. While datasets for autonomous vehicles are abundant, there is a critical gap in datasets tailored for outdoor human navigation. This gap poses a major obstacle to the development of computer vision based \textit{Assistive Technologies}. To overcome this obstacle, we present SANPO, a large-scale egocentric video dataset designed for dense prediction in outdoor human navigation environments. SANPO contains 701 stereo videos of 30+ seconds captured in diverse real-world outdoor environments across four geographic locations in the USA. Every frame has a high resolution depth map and 112K frames were annotated with \textit{temporally consistent} dense video panoptic segmentation labels. The dataset also includes 1961 high-quality synthetic videos with pixel accurate depth and panoptic segmentation annotations to balance the noisy real world annotations with the high precision synthetic annotations.

SANPO is already publicly available and is being used by mobile applications like Project Guideline to train mobile models that help low-vision users go running outdoors independently. To preserve anonymization during peer review, we will provide a link to our dataset upon acceptance.


\end{abstract}
    \section{Introduction}
\label{section:introduction}
\begin{figure}[t]%
    \centering
    \subfloat{{\includegraphics[width=1\linewidth]{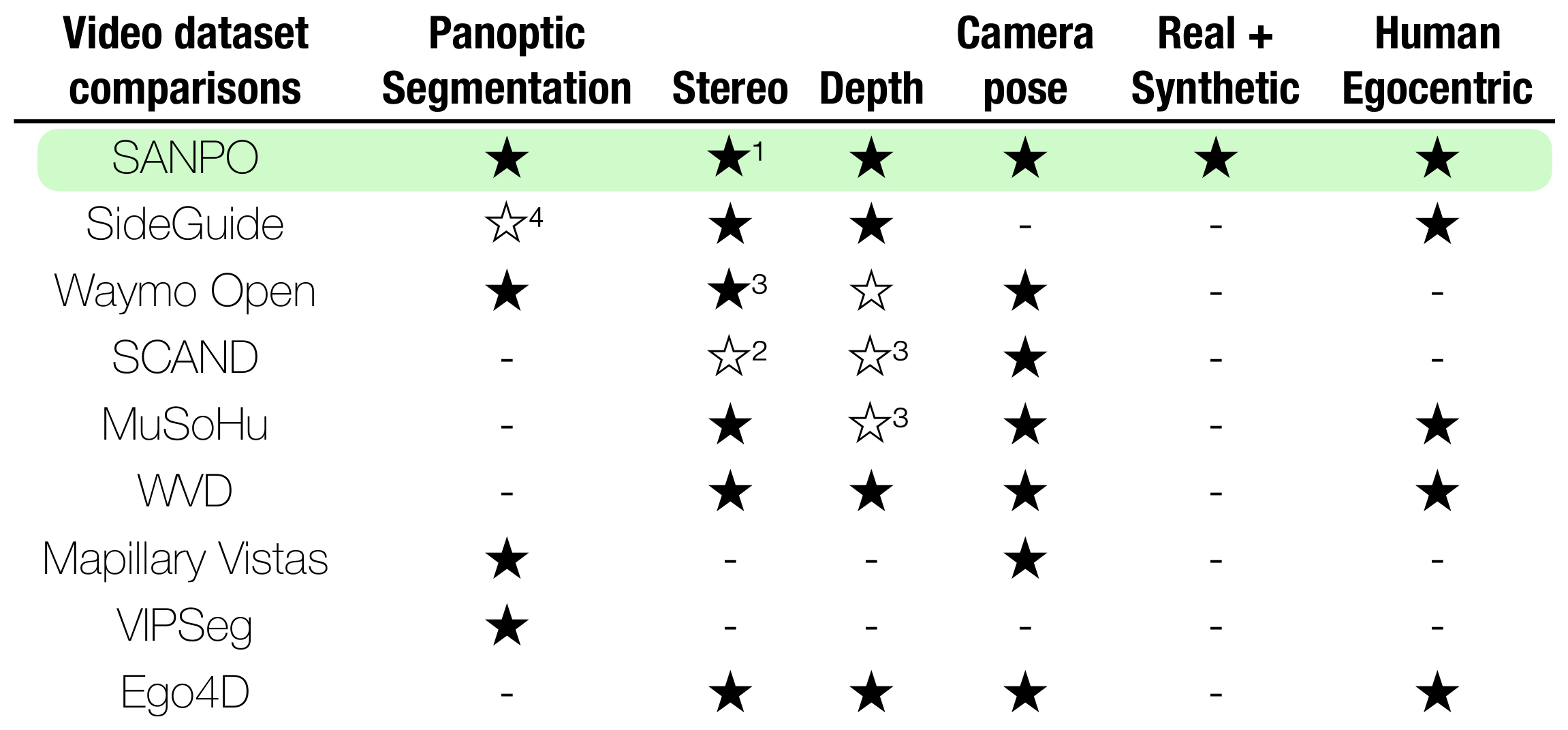}}}%
    \\
    \subfloat{{\includegraphics[width=1\linewidth]{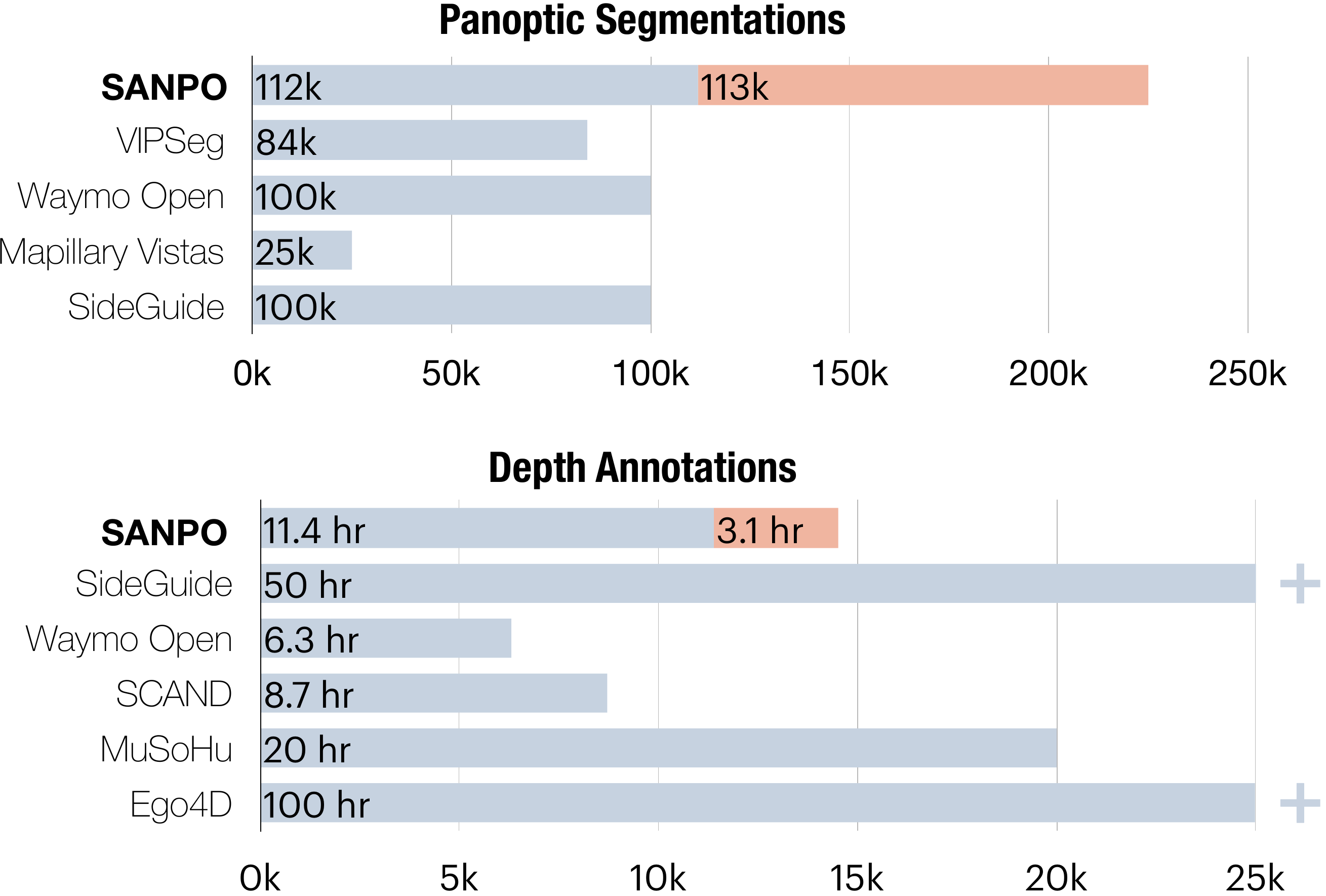}}}%
    \caption{\label{fig:sanpo-compare}\textbf{SANPO} is the only human-egocentric dataset with panoptic masks, multi-view stereo, depth, camera pose, and both real and synthetic data. SANPO has the largest number of panoptic frames among related work and a respectable number of depth annotations. (Note: $^1$: multi-view, $^2$: partial coverage, $^3$: sparse depth, $^4$: sparse segmentation)}
\end{figure}
An estimated 295 million people worldwide experience moderate to severe visual impairment \cite{bourne2021trends}. Advances in human navigation systems \cite{kuriakose2022tools} could significantly improve their quality of life. To realize this potential, AI systems require robust egocentric scene understanding.  Building this capability depends on the availability of large-scale, high-quality, and diverse datasets.
\begin{figure*}[t]%
\centering
    \subfloat{{\includegraphics[bb=0 0 46cm 24cm,width=.2666\linewidth]{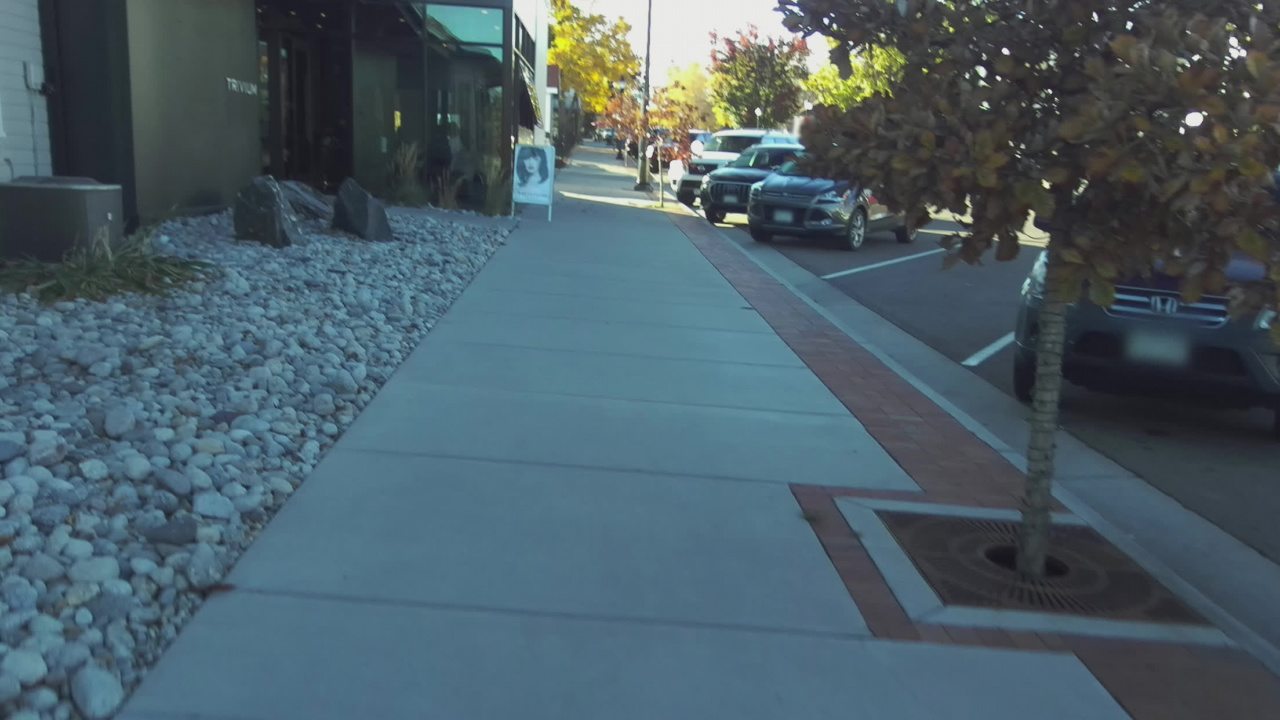}}}%
    \subfloat{{\includegraphics[bb=0 0 46cm 24cm,width=.2666\linewidth]{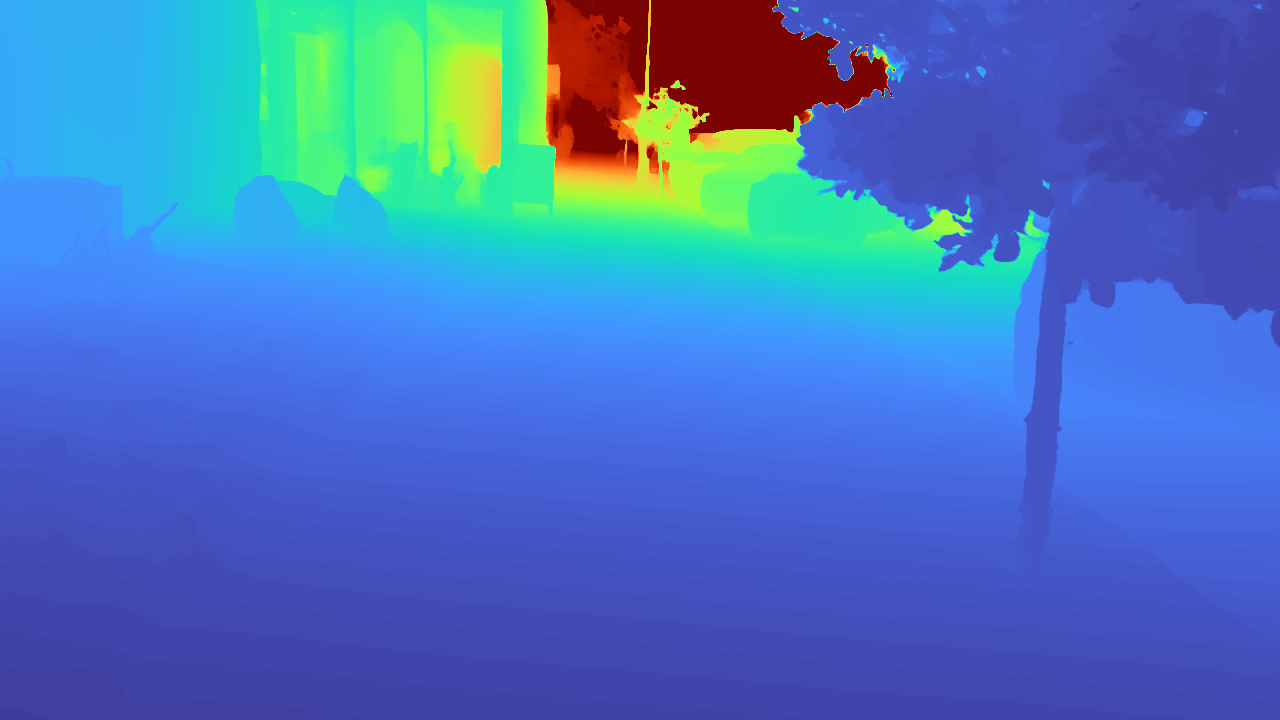}}}%
    \subfloat{{\includegraphics[bb=0 0 46cm 24cm, width=.2666\linewidth]{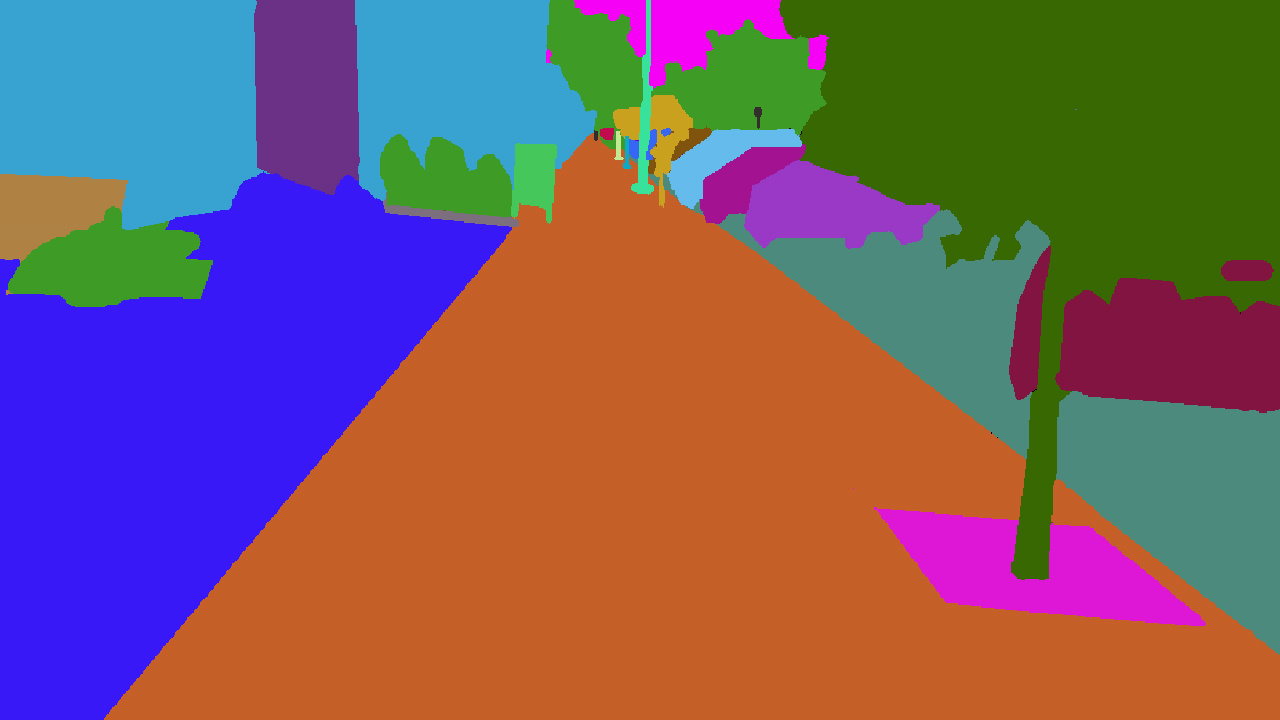}}}%
    
    
    \subfloat{{\includegraphics[width=0.8\linewidth, trim={0 10cm 0 10cm},clip]{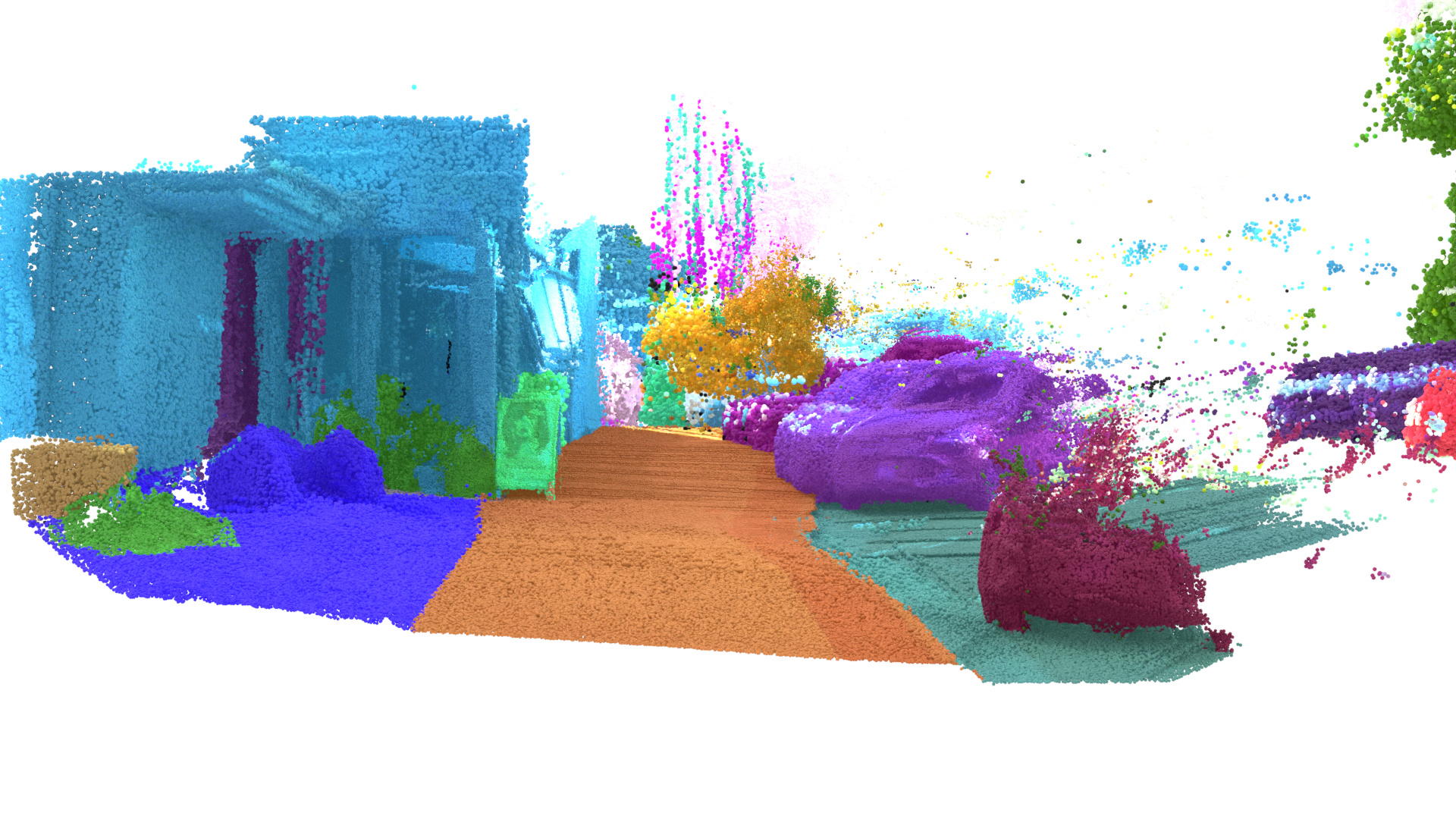}}}%
    \caption{\textbf{SANPO Real Sample.} Top row shows a stereo left frame from a session along with its metric depth and segmentation annotations. Bottom row shows the 3D scene of the session built using the annotations we provide. Points from several seconds of video are accumulated and aligned with ICP.}%
    \label{fig:sanpo-sample-real}%
    \centering
\end{figure*}
The past decade has seen a surge in datasets for autonomous driving \cite{kang2019test, Mao_2022_ACCV, wilson2023argoverse, mei2022waymo}. However, these datasets are not suitable for the unique challenges of egocentric human navigation because they are are designed for self-driving vehicles.
These challenges include unusual viewpoints, motion artifacts, and dynamic human-object interactions. Unlike cars, which primarily operate in structured environments, humans navigate spaces that are often cluttered, unpredictable, and less regulated.
This critical lack of publicly available human navigation-specific datasets hinders the development of robust assistive technologies.

To address this, we present SANPO, a comprehensive dataset designed to support research on outdoor human egocentric scene understanding. SANPO offers a rich combination of real and synthetic data. It is comprised of two underlying datasets: \textbf{SANPO-Real} contains 617K stereo video frame pairs with 617K estimated depth maps and 112K dense video panoptic masks. \textbf{SANPO-Synthetic} contains 113K video frames with pixel-accurate depth maps and dense video panoptic masks.

SANPO's diversity and complexity also makes it an invaluable resource for advancing dense prediction tasks beyond human navigation. The inclusion of two stereo camera videos for each real session offers further potential for advancing multi-view methods.
    
\section{Related Work}
\label{sec:related-work}
We can categorize the relevant scene understanding datasets into four categories: (i) Accessibility and Human Navigation, (ii) Human Egocentric, (iii) Autonomous Driving, and (iv) General Scene Understanding.
See~\ref{appendix:dataset_comparison} in the supplementary material for a comparison of many contemporary datasets.

\subsection{Accessibility and Human Navigation}
Few datasets in the literature explore accessibility and human navigation \cite{9340734, Neuhold_2017_ICCV, ahmed2017optimization, yang2019robustifying}, with SideGuide \cite{9340734} standing out as the most relevant and extensive 
example.  While both SANPO and SideGuide share a focus on egocentric human navigation, SANPO differentiates itself through several key advantages:
i) Scale and Diversity: SANPO-Real surpasses SideGuide with a significantly larger dataset (617K stereo pairs vs. 180K) and broader environmental diversity, spanning urban, suburban, parks, trails, open terrain, etc. compared to SideGuide's focus on urban sidewalks. 
ii) Temporally Consistent Video Segmentation Annotation: SANPO provides dense, temporally consistent video panoptic segmentation annotations, whereas, SideGuide offers sparse object masks. 
iii) Depth Labels: While both datasets provide estimated depth labels, SANPO utilizes CREStereo \cite{li2022practical} for potentially higher accuracy. 
iv) Synthetic Counterpart: SANPO includes a synthetic counterpart designed specifically to empower synthetic-to-real domain adaptation research.


\subsection{Human Egocentric}
Several datasets provide relevant data from a human egocentric perspective. SCAND \cite{karnan2022socially}, designed for robot navigation, includes depth and odometry labels but lacks the semantic segmentation crucial for scene understanding in human navigation. MuSoHu \cite{nguyen2023toward}, captured with various sensors including a stereo camera, LIDAR, microphone array, and a 360-degree camera, offers depth and odometry labels while exhibiting motion artifacts typical of human movement. However, it also lacks semantic segmentation labels. Ego4D \cite{grauman2022ego4d} is a large-scale dataset, but its focus extends beyond human navigation, and it lacks the necessary semantic segmentation for this task.

On the synthetic front, MOTSynth \cite{fabbri21iccv} and WVD \cite{nadeem2019wvd} offer limited relevance. MOTSynth, while including segmentation and depth annotations, is limited by its focus on pedestrian-only annotations and restricted egocentric views. WVD, designed for violence detection, lacks both segmentation and depth labels essential for understanding the navigation environment.

\subsection{Autonomous Driving}
Datasets for autonomous vehicles, such as Cityscapes \cite{Cordts2016Cityscapes, vip_deeplab}, Argoverse \cite{wilson2023argoverse}, Waymo \cite{mei2022waymo}, and CamVid \cite{BROSTOW200988}, among others \cite{Mao_2022_ACCV, Liao2022PAMI, richter2017playing, nuscenes2019, astar-3d}, are prevalent in the field of computer vision. While these datasets often include rich annotations like stereo video, segmentation, depth, and 3D object detection labels, they prioritize vehicle-centric perception.
Though there are several such datasets, they are not sufficient to address the unique challenges posed by human navigation environments. Egocentric environments often feature unstructured, unpredictable, and cluttered scenes, with unusual viewpoints, and motion artifacts due to human movement. They also exhibit dynamic human-object interactions. Addressing these challenges requires specialized datasets tailored to human navigation scenarios.

\subsection{General-Purpose Scene Understanding}
Datasets such as MSCOCO \cite{lin2014microsoft}, DAVIS-2017 \cite{Caelles_arXiv_2019}, and YouTube-VOS \cite{xu2018youtube} focus on object detection and segmentation. Similar to autonomous driving datasets, they offer rich annotations but lack the attributes required for building a human navigation system.

Overall, existing datasets, though plentiful,
don't adequately satisfy the requirements of human navigation in diverse environments. See~\ref{appendix:dataset_comparison} in the supplementary material for a broad comparison.
\begin{figure}[t]
\includegraphics[width=1\linewidth]{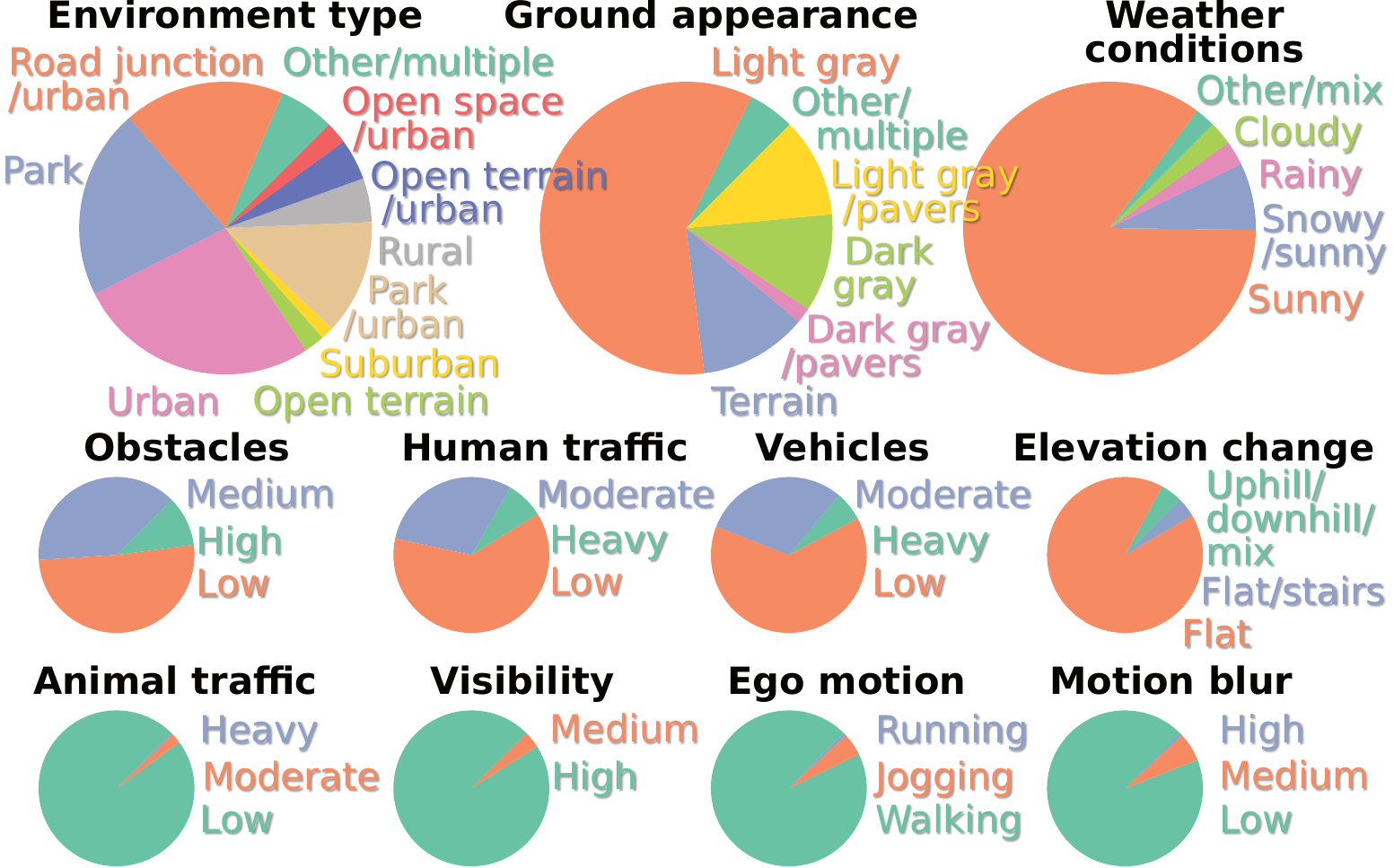}
\caption{\label{fig:annotation-distributions}\textbf{SANPO-Real environment diversity,} showing the distribution of video-level annotations for 11 of the 12 attributes. Each pie chart shows that annotation over all 701 sessions.}
\end{figure}
    \begin{figure*}[t]%
    \centering
    \begin{minipage}{0.64\linewidth}
    \subfloat{{\includegraphics[bb=0 0 52cm 28cm,width=1\linewidth,trim=0 2cm 0 5cm,clip]{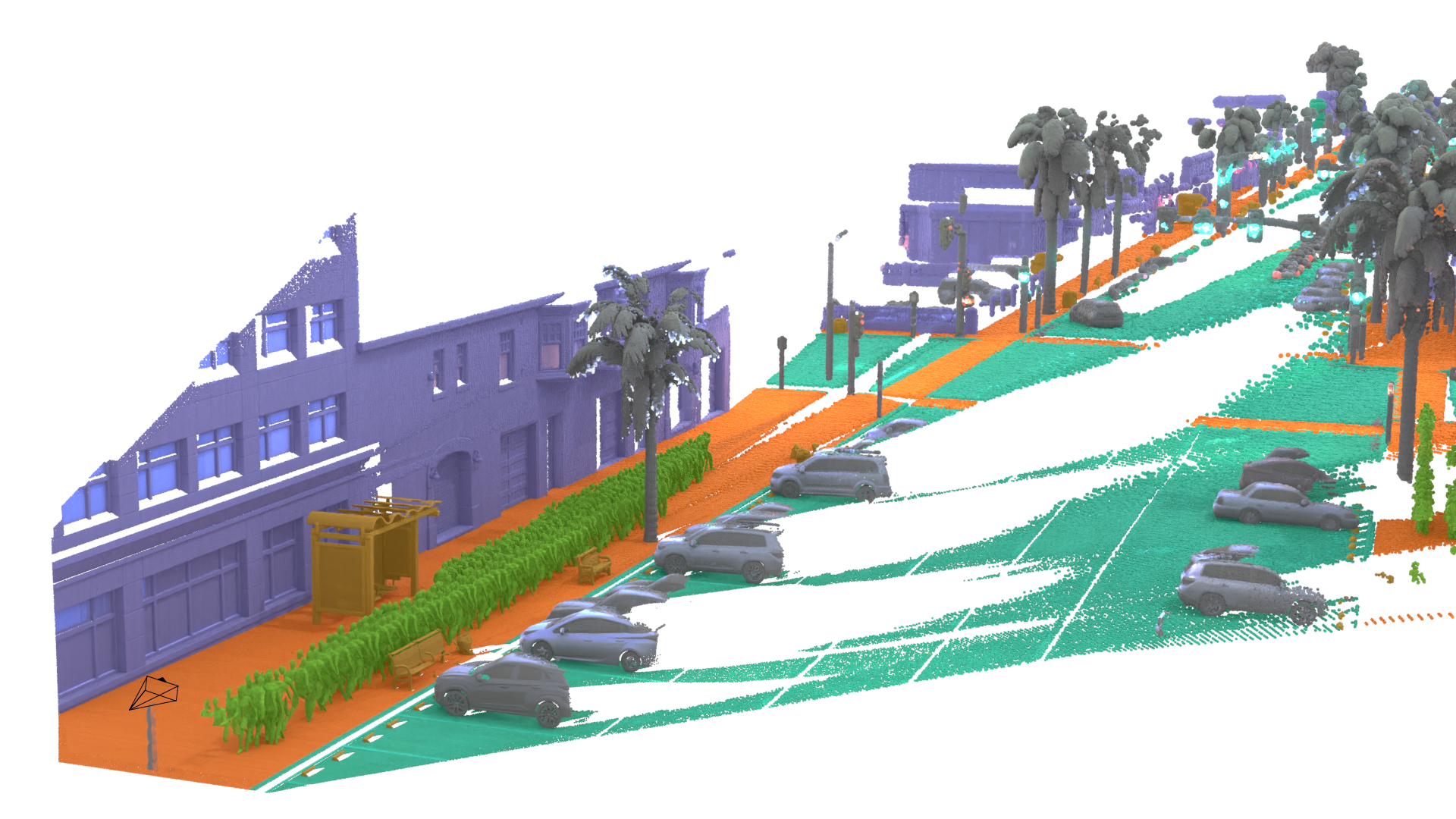}}}%
    \end{minipage}
    \begin{minipage}{0.16\linewidth}
    \subfloat{{\includegraphics[bb=0 0 80cm 44cm,width=1\linewidth]{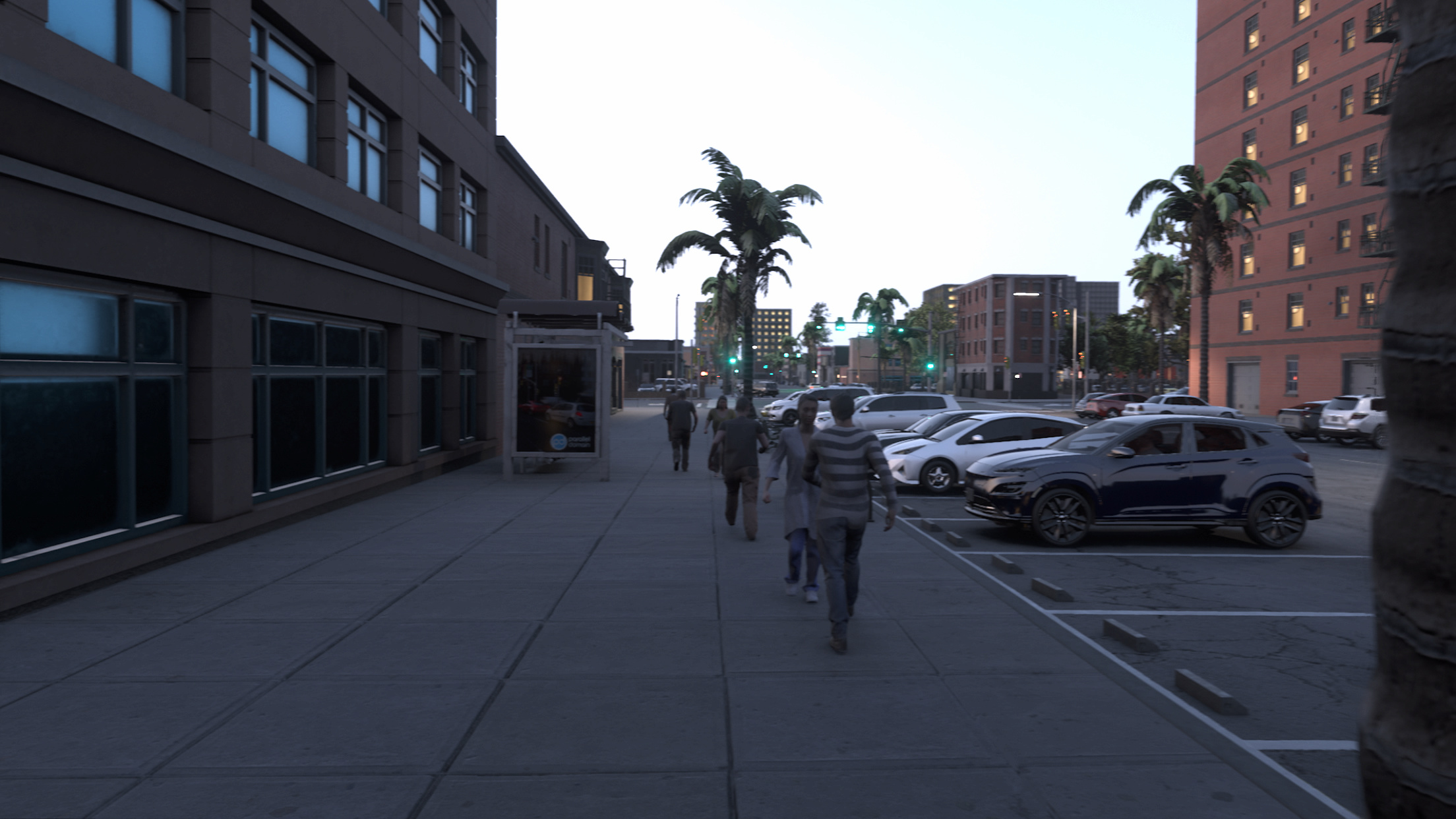}}}%
    
    \subfloat{{\includegraphics[bb=0 0 80cm 44cm,width=1\linewidth]{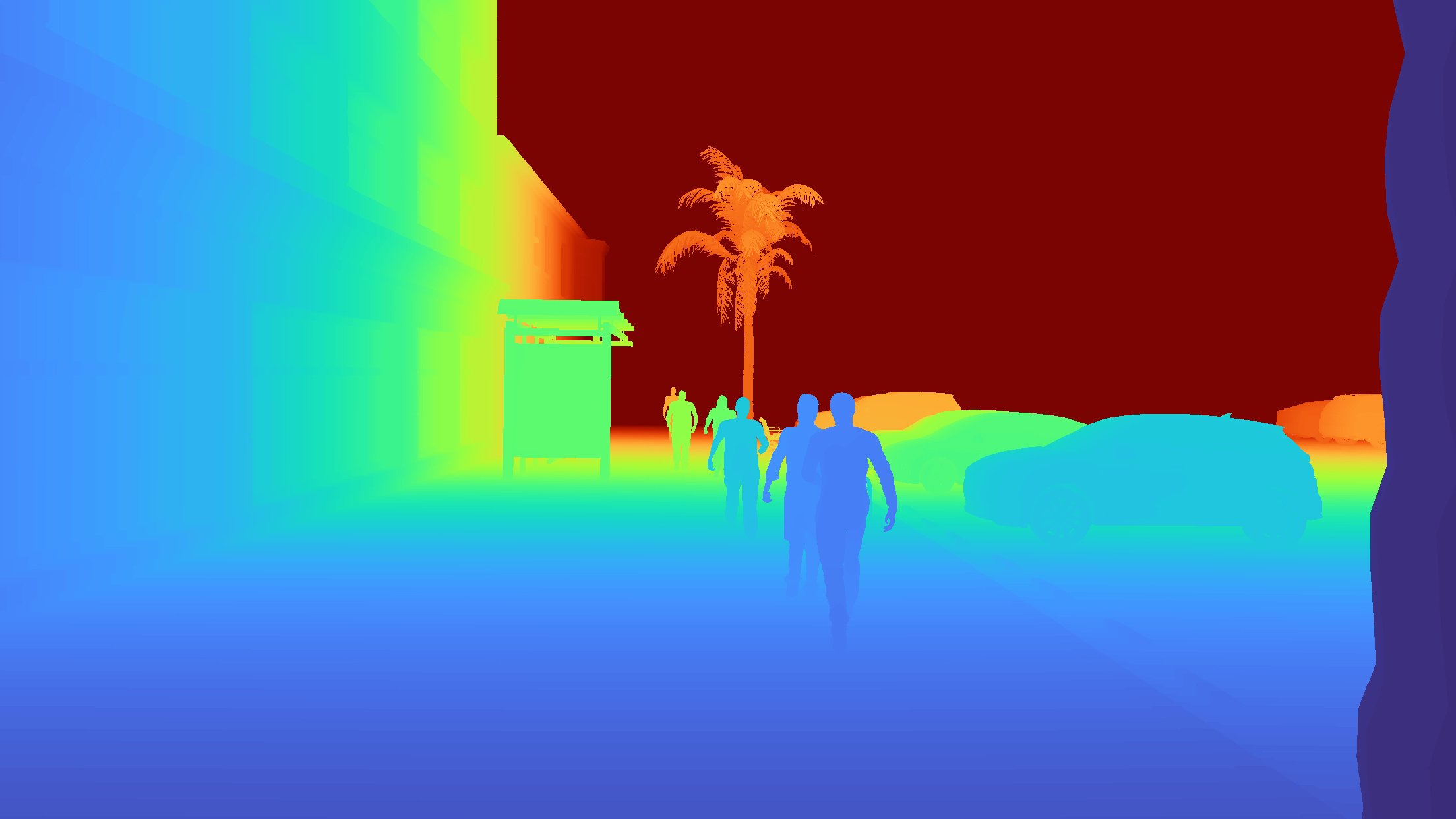}}}%
    
    \subfloat{{\includegraphics[bb=0 0 80cm 44cm,width=1\linewidth]{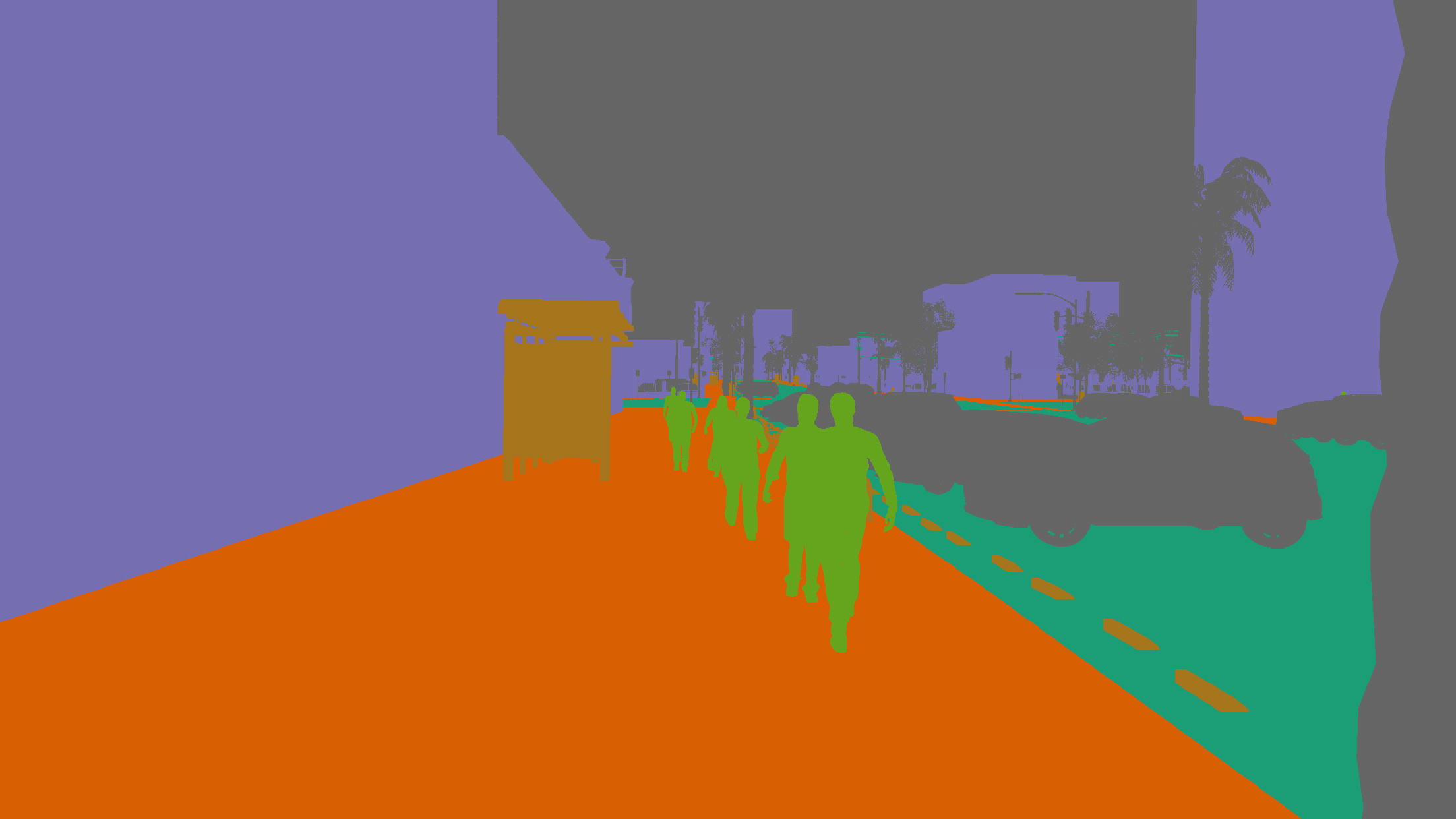}}}%
    \end{minipage}
    \caption{\textbf{SANPO-Synthetic Sample.} Right column shows a single frame from a synthetic session along with its metric depth and segmentation annotation. Left column shows the 3D scene of the session built using the annotations. Points come from the accumulated depth maps and camera locations across many frames.}%
    \label{fig:sanpo-sample-synthetic}%
\end{figure*}
\section{SANPO}
SANPO is comprised of
\textit{SANPO-Real}, a collection captured by real-world volunteer runners, and \textit{SANPO-Synthetic}, a ``digital twin'' recreation in a virtual environment. In this section, we provide a detailed overview of both. 

\subsection{SANPO-Real}
\label{sec:sanpo-real-dataset}


\subsubsection{Data Collection}
\label{section:data-collection}


\paragraph{Collection Statistics}
SANPO-Real consists of 701 sessions recorded simultaneously with two stereo cameras, yielding four RGB video streams per session. Each video is approximately 30 seconds long, captured at 15 frames per second (FPS). We provide all videos in a lossless format to support stereo vision research, with 597 sessions at a resolution of $2208\times1242$ pixels and the remainder at $1920\times1080$ pixels.
All videos were rectified using ZED software.
In total, we collected 617,408 stereo frames. The chest-mounted ZED-2i, equipped with a 4mm lens, captured 308,957 stereo frames. This setup provided long-range depth from a stable mounting point, minimizing motion blur. The lightweight head-mounted ZED-M contributed 308,451 stereo frames, offering a wide field of view for comprehensive video.

\paragraph{Camera rig}
To accommodate SANPO-Real's multi stereo camera requirements, we designed a specialized data collection rig for our volunteer runners to wear.
Please see \ref{appendix:rig} in the supplementary material to learn more about
it.

\paragraph{Ethics, Diversity and Privacy}
A dedicated team of volunteers employed at our company
meticulously collected data. To ensure integrity, we provided clear guidelines on where and how to collect data, and strictly adhered to all relevant local, state, and city laws.

We collected data across four diverse US locations: San Francisco CA, Mountain View CA, Boulder CO, and NYC. These regions span urban areas, suburbs, city streets, and public parks. Our volunteers captured data under a wide range of conditions within each region, varying weather (including snow and rain), different times of day%
, and diverse ground types. They also navigated obstacles, varied their speeds, and encountered varying levels of traffic.

First, each volunteer reviewed their collected samples, immediately deleting any that didn't meet our guidelines. Next, all the remaining videos were processed to remove personally identifiable information (PII), specifically faces and license plates
\footnote{We computed depth annotations before blurring PII, as blurry patches can lead to inaccurate results.}. Finally, only after completing these privacy measures did we upload the videos to our cloud storage.

\subsubsection{Annotation}
Each session includes session-level attributes such as human traffic, density of obstacles, environment type \textit{etc,}\ 12 in total.
The distribution of these annotations in Fig.~\ref{fig:annotation-distributions} showcases the diversity of environments present in SANPO-Real.
Additionally, we provide camera poses for every stereo recording derived from fused IMU and VIO measurements provided by the ZED software. 
Finally, we provide depth and panoptic segmentation annotations.

\paragraph{Depth}
For each stereo video we provide a sparse depth map generated by the ZED SDK and, similar to SideGuide \cite{9340734}, a dense depth map estimated using a stereo algorithm.
We use CREStereo~\cite{li2022practical}, a state-of-the-art ML-based stereo depth model.
CREStereo maps stereo frames to a disparity map, which we convert to depth using camera intrinsics and limit the range to 0-80 meters.  These depth maps have resolution of 1280$\times$720 pixels, a result of CREstereo's architecture.

\paragraph{Temporally Consistent Panoptic Segmentation}
We developed a segmentation taxonomy specifically for egocentric human navigation.
This taxonomy balances panoptic segmentation with the need for detailed analysis of the navigation environment. It includes 31 categories: 15 "\emph{thing}" classes (distinguishable objects) and 16 "\emph{stuff}" classes (amorphous regions). A detailed taxonomy
is provided in \ref{appendix:sanpo-taxonomy} in the supplementary material.
Using this taxonomy we provide panoptic segmentation annotations for 237 sessions. We annotated the left stereo video from one camera within each session, resulting in annotations for 146 videos captured by the long-range ZED-2i camera and 91 videos captured by the wide-angle ZED-M camera. All annotations are temporally consistent, ensuring that each object maintains a unique ID throughout the video.
SANPO-Real offers a combination of human-annotated and machine-propagated annotations. 18,787 human-annotated and 93,981 machine-propagated frames. A total of 975,207 object mask, with 195,187 manually segmented and 780,020 machine propagated.
Section \ref{appendix:seg_process} in the supplementary material details the annotation process.
Figure~\ref{fig:sanpo-sample-real} provides a visual example of a SANPO-Real session.

\subsection{SANPO-Synthetic}
\label{sec:sanpo-synthetic-dataset}

The high cost of building large-scale real-world datasets~\cite{peng2018visda, reddy2023synthetic}
creates significant interest in synthetic-to-real domain adaptation research~\cite{peng2018visda, reddy2023synthetic}.
To support this research and complement SANPO-Real, we partnered with \emph{Parallel Domain} to generate high-quality synthetic data. We worked closely with \emph{Parallel Domain} to define simulation environments, providing detailed specifications and distributions for various simulation entities. The synthetic environment was optimized to replicate real-world capture conditions, including scenery, camera parameters, object placement and frequency. Through an iterative process, we achieved a synthetic dataset that closely matches SANPO-Real.
Please see section \ref{appendix:sanpo-synthetic-reproducibility} in the supplementary material for full specifications of the rendering environment (camera intrinsics, obstacle distribution, \textit{etc.})

\subsubsection{Simulation Statistics and Annotations}
SANPO-Synthetic comprises 113,794 annotated monocular video frames from 1961 rendered sessions, simulating real-world camera configurations. To achieve this, 960 sessions utilize a (simulated) chest-level ZED-2i, while 1001 sessions employ a head-mounted ZED-M, with parameters mirroring their physical counterparts. This dataset features diverse frame rates of 5, 14.28, and 33.33 FPS.

Each synthetic video is accompanied by camera pose trajectories, dense depth maps and temporally consistent panoptic segmentation masks. A key attribute of SANPO-Synthetic is its pixel-perfect annotations.
Figure~\ref{fig:sanpo-sample-synthetic} provides a visual example of a SANPO-Synthetic session.

    \begin{figure*}[h]
\centering
\includegraphics[width=0.8\linewidth]{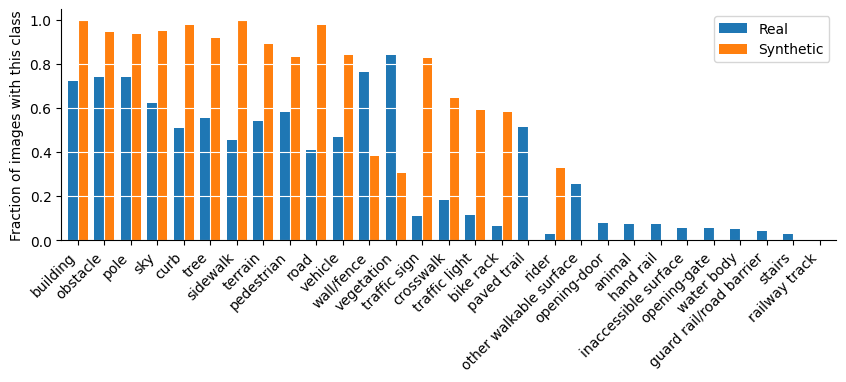}
\caption{\label{fig:class-hist}
\textbf{Semantic label occurrences in SANPO}: Common human navigation specific labels like Building, Obstacle, Pole, Tree, Curb, Sidewalk etc. feature more prominently.  
}
\end{figure*}

\section{Analysis}
\begin{figure}[t]
\centering

    \begin{minipage}{0.49\linewidth}
    \subfloat{{\includegraphics[width=1\linewidth]{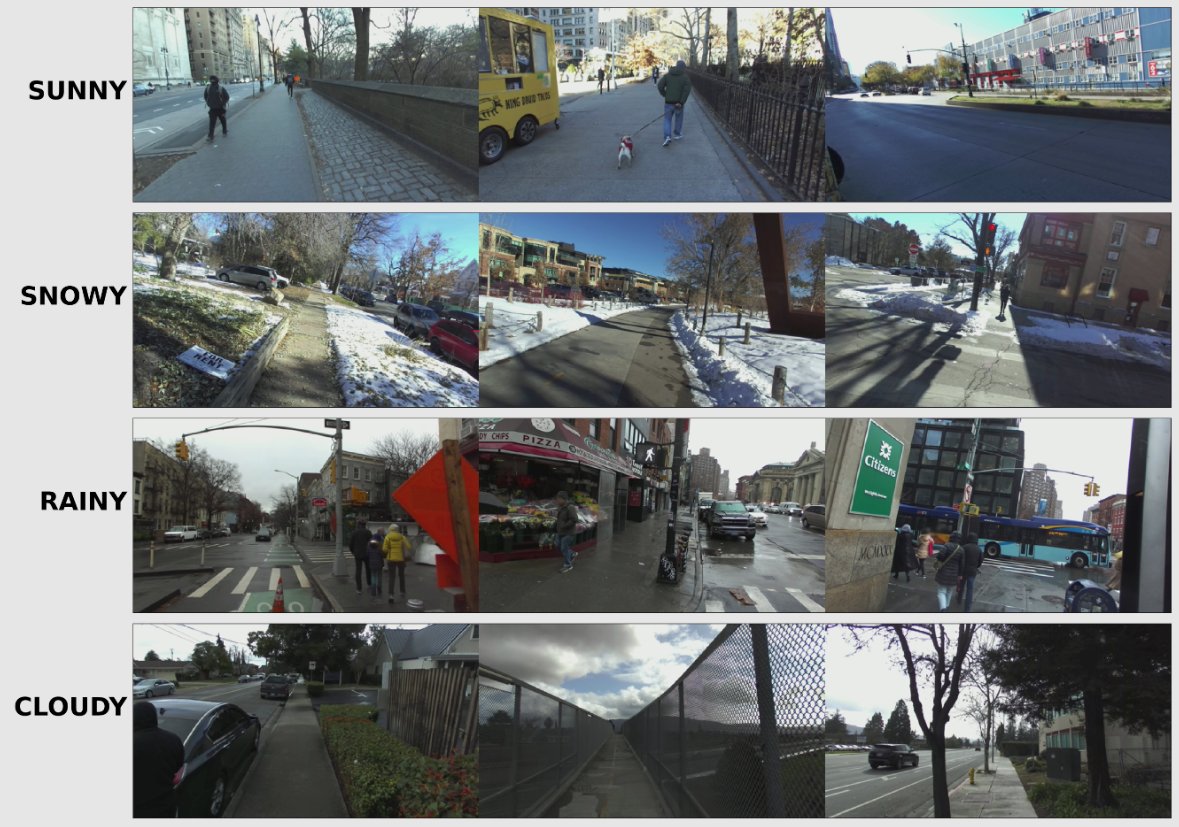}}}%
    \end{minipage}
    \begin{minipage}{0.49\linewidth}
    \subfloat{{\includegraphics[width=1\linewidth]{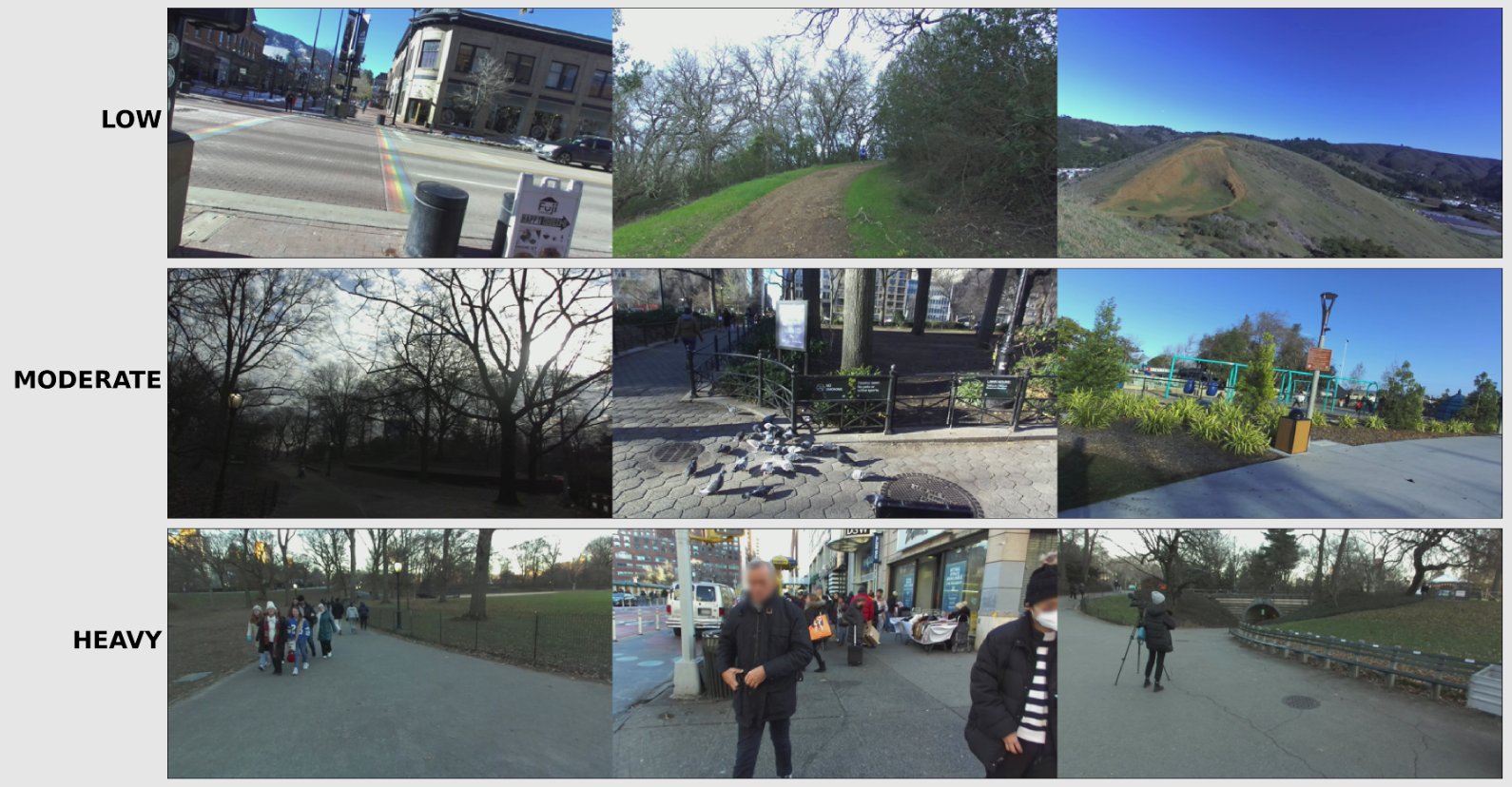}}}%
    \end{minipage}

\caption{\label{fig:attribute-weather}
\textbf{SANPO is a Diverse Dataset.} An example of SANPO's data diversity across weather conditions~(left) and human traffic levels~(right) based on the session-level attributes' annotation. Please zoom in for better view.
}

\end{figure}

SANPO offers a comprehensive dataset of 2662 sessions, encompassing 701 real sessions (``SANPO-Real'') captured across diverse real-world environments and conditions, and 1961 synthetic sessions (``SANPO-Synthetic'') meticulously simulated with varying distributions of simulation entities.


\paragraph{Data Diversity}
Figure~\ref{fig:attribute-weather} demonstrates the breadth of SANPO's data diversity by showcasing real samples across various weather conditions and human traffic levels, categorized based on session-level attributes' annotation.
These session-level attributes could potentially be used to create views of SANPO to satisfy applications' data requirements.

\paragraph{Distribution of Semantic Labels and Instances}
Figure~\ref{fig:class-hist} illustrates the distribution of semantic labels as a percentage of images containing them. Common human navigation specific labels like Building, Obstacle, Pole, Tree, Curb, Sidewalk, \textit{etc.}\ are more prevalent.
The quantity of annotated objects across the dataset is shown in Figure~\ref{fig:instance-hist}. 
The density of pedestrians is shown in the supplementary material; see~Fig.~\ref{fig:pedestrians} in \ref{ref:additional-statistics}.
Due to the controlled nature of virtual environments—where each object is masked including the objects distant from the egoperson—SANPO-Synthetic exhibits a considerably higher number of instances compared to SANPO-Real.
\paragraph{Depth Distribution}
Figure~\ref{fig:sanpo-depth-dist} shows the depth distribution of annotated objects within the dataset, revealing a contrast between SANPO-Real and SANPO-Synthetic. Both datasets have many annotated objects further than 20m away, but the distribution of SANPO-Synthetic skews much further than SANPO-Real.
This is because the synthetic rendering pipeline can generate segmentations for objects that are arbitrarily far away.


\begin{figure}[t]
\centering
\includegraphics[width=1\linewidth]{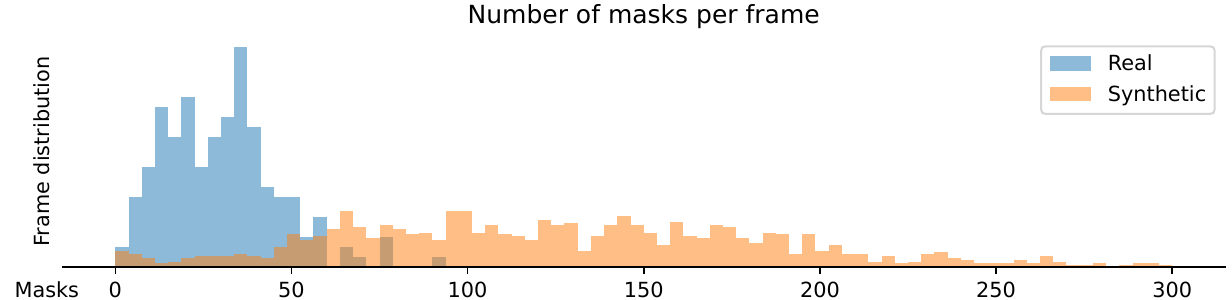}
\caption{\textbf{Mask Quantity in SANPO}. A normalized histogram of ``object'' counts in SANPO, where  ``objects'' are connected components of each unique panoptic label covering at least 100px.
Real-world images typically have fewer annotated instances per image, rarely exceeding 40. In contrast, synthetic images feature significantly higher instance density, reflecting the controlled nature of virtual environments where each object can be individually masked.}
\label{fig:instance-hist}
\end{figure}
\begin{figure}[t]
\centering
\includegraphics[width=1\linewidth]{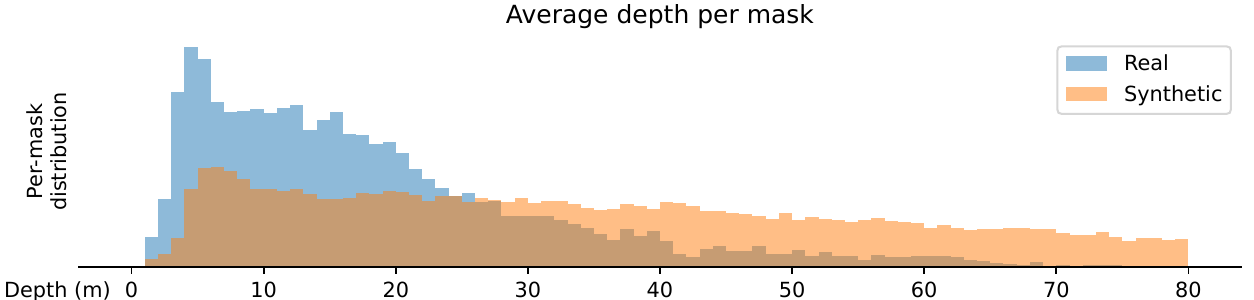}
\caption{\textbf{SANPO Depth Distribution.} A normalized histogram of the average depth of each annotated object in SANPO-Real and SANPO-Synthetic, where ``annotated objects'' are connected components of each unique panoptic label covering at least 100px.
Objects in SANPO-Synthetic tend to be further away than SANPO-Real because of the pixel-perfect segmentation and depth groundtruth. In both datasets, there are many objects further than 20m away, creating interesting challenges for obstacle detection applications.
}
\label{fig:sanpo-depth-dist}
\end{figure}
    \begin{table*}

\centering

\begin{tabular}{ccccc}
\toprule
 \multicolumn{5}{c}{\textbf{Pretrained Semantic Segmentation Model on Cityscape \cite{Cordts2016Cityscapes}}}
 \\\cmidrule(lr){1-5}
 Method & Encoder & \begin{tabular}{@{}c@{}}mIoU $\uparrow$ \\ SANPO-Real\end{tabular} & & \begin{tabular}{@{}c@{}}mIoU $\uparrow$ \\
 Cityscapes \cite{Cordts2016Cityscapes}\end{tabular}
 \\\cmidrule(lr){1-5}
 Kmax-Deeplab \cite{yu2023kmaxdeeplab} & ResNet-50 \cite{he2016deep} &  38.6 & & 79.7
 \\ 
 Kmax-Deeplab \cite{yu2023kmaxdeeplab} & ConvNeXt-L \cite{liu2022convnet} & 42.4 & & 83.5
 \\
 
 \midrule
 
 \multicolumn{5}{c}{\textbf{Segmentation Foundation Model}}
 \\\cmidrule(lr){1-5}
 Method & Model & \begin{tabular}{@{}c@{}}Instance mIoU $\uparrow$ \\
 SANPO-Real\end{tabular} & & \begin{tabular}{@{}c@{}}Instance mIoU $\uparrow$ \\ Cityscapes \cite{Cordts2016Cityscapes}\end{tabular}
 \\\cmidrule(lr){1-5}
 Center Point Prompt \cite{kirillov2023segment} & SAM \cite{kirillov2023segment} & 48.9 & & 41.0
 \\
 \midrule
 
 \multicolumn{5}{c}{\textbf{Pretrained Depth Estimation Model}}
 \\\cmidrule(lr){1-5}
 Method & Encoder & \begin{tabular}{@{}c@{}}$\depthmetric~\uparrow$ \\ SANPO-Real\end{tabular} & \begin{tabular}{@{}c@{}}$\depthmetric~\uparrow$ \\ KITTI \cite{Geiger2012CVPR}\end{tabular} &
 \begin{tabular}{@{}c@{}}$\depthmetric~\uparrow$ \\ Cityscapes-DVPS \cite{vip_deeplab}\end{tabular}
 \\\cmidrule(lr){1-5}
 ZoeDepth \cite{bhat2023zoedepth} & BEiT$_{384}$-L \cite{DBLP:journals/corr/abs-2106-08254} & 0.135 & 0.97 & 0.39\\
 
 \midrule
 
 \multicolumn{5}{c}{\textbf{Depth Foundation Model}}
 \\\cmidrule(lr){1-5}
 Method & Model & \begin{tabular}{@{}c@{}}$\depthmetric~\uparrow$ \\ SANPO-Real\end{tabular} & \begin{tabular}{@{}c@{}}$\depthmetric~\uparrow$ \\ KITTI \cite{Geiger2012CVPR}\end{tabular} &
 \begin{tabular}{@{}c@{}}$\depthmetric~\uparrow$ \\ Cityscapes-DVPS \cite{vip_deeplab}\end{tabular}
 \\\cmidrule(lr){1-5}
 Metric Depth & Depth-Anything \cite{depthanything} & 0.22 & 0.98 & 0.57
 \\
  
 \bottomrule
\end{tabular}
\caption{\textbf{SANPO is a Challenging Dataset.} Zero-shot performance of various pre-trained models on the SANPO-Real test set. Higher values indicate better performance across all metrics.}
\label{table:zero-shot}
\end{table*}
\section{Benchmarks}
\label{sec:benchmarks}
In this section, we establish baselines for SANPO dataset in two distinct evaluation settings: 1) \textbf{Zero-Shot Evaluation}: We assess the ability of published model checkpoints to generalize directly to the SANPO dataset. 2) \textbf{SANPO Benchmark}: We establish baselines for state-of-the-art architectures on dense prediction tasks using the SANPO dataset.
Please see~\ref{appendix:sanpo-benchmark-setup} in the supplementary material for detailed experimental setup.
    
\paragraph{Metrics}
We evaluate the models' performance using the following well established metrics:
\begin{itemize}
\itemsep-0.2em 
    \item Semantic Segmentation: Mean Intersection over Union (mIoU) as in \cite{yu2023kmaxdeeplab}.
    \item Panoptic Segmentation: Panoptic Quality (PQ) as in \cite{yu2023kmaxdeeplab}.
    \item Depth Estimation: Fraction of depth inliers across the depth map, $\mathbb{E}\left[\max(\frac{y}{y'}, \frac{y'}{y}) \leq 1.25\right]$, denoted as \depthmetric{}, as in \cite{bhat2023zoedepth}. 
\end{itemize}
We report metrics on the test set of SANPO-Real. For semantic and panoptic segmentation, metrics are reported only on the human-annotated subset of the SANPO-Real test set.

\subsection{Zero-Shot Evaluation}
Our goal with SANPO is to provide a dataset representative of outdoor human navigation tasks from an egocentric perspective, which is distinct from domains like autonomous driving. This evaluation establishes zero-shot baselines and assesses the dataset's challenge for zero-shot prediction. 

\paragraph{Semantic Segmentation Baseline} For this baseline we used the state-of-the-art Kmax-Deeplab~\cite{yu2023kmaxdeeplab} trained on the Cityscapes dataset~\cite{Cordts2016Cityscapes}. For a fair comparison, we mapped the Cityscapes taxonomy to SANPO taxonomy, excluding $18$ SANPO classes that lack direct mapping. We do not report panoptic quality for this baseline due to mismatch in \emph{"thing"} classes of the Cityscapes and SANPO. For details on the Cityscapes to SANPO taxonomy mapping please refer to~\ref{appendix:cityscapes-to-sanpo}
in the supplementary material.

\paragraph{Depth Estimation Baseline}
For this baseline, we used the publicly available checkpoint for ZoeDepth-M12~NK~\cite{bhat2023zoedepth}. 


\paragraph{Foundation Models Baseline} We also evaluated two foundation models: 1) \textbf{Segment Anything Model (SAM) \cite{kirillov2023segment}}, using the center point prompt and reporting instance-level mIoU for interactive segmentation tasks \cite{ritm}. We excluded instances smaller than 2\% of the image size for evaluation efficiency. 2) \textbf{Depth-Anything \cite{depthanything}}, using their outdoor metric depth model.

\cut{\begin{enumerate}
    \item Semantic Segmentation:  Segment Anything Model (SAM) \cite{kirillov2023segment}, using the center point prompt and reporting instance-level mIoU for interactive segmentation tasks \cite{ritm}. We excluded instances smaller than 2\% of the image size for evaluation efficiency.
    
    \item Depth Estimation:  Depth-Anything \cite{depthanything}, using their outdoor metric depth model.
\end{enumerate}}

Our findings (Table~\ref{table:zero-shot}) highlight the challenge SANPO presents for depth and segmentation models. 
Despite their focus on metric depth estimation and zero-shot transfer, ZoeDepth~\cite{bhat2023zoedepth} and Depth-Anything~\cite{depthanything} 
demonstrate limited performance on SANPO-Real with \depthmetric{} scores of 0.135 and 0.22, respectively. This underscores the need for more diverse metric depth datasets to improve generalization.
Similarly, for segmentation, Kmax-Deeplab (ConvNeXt-L) exhibits a dramatic performance drop from its 83.5 mIoU on Cityscapes to 42.4 on SANPO-Real (using human-annotated ground truth only).
This demonstrates that SANPO is generally a quite challenging dataset for current state-of-the-art segmentation methods.
\subsection{SANPO Benchmark}
\begin{table}[t]
{

\centering
\scalebox{0.87}{
\begin{tabular}{cccc}
\toprule

\multicolumn{4}{c}{\textbf{Panoptic Segmentation : Kmax-DeepLab \cite{yu2023kmaxdeeplab}}}\\
\midrule

\textbf{Dataset} & \textbf{Encoder} 
& \textbf{mIoU $\uparrow$}
& \textbf{PQ $\uparrow$}\\
\midrule

SANPO-Synthetic-Train &  Resnet-50 \cite{he2016deep}	& 14.6	& 9.6\\
SANPO-Real-Train &  Resnet-50 \cite{he2016deep} & 43.4 & 34.6\\

SANPO-Synthetic-Train & ConvNeXt-L \cite{liu2022convnet} & 17.6	& 11.1\\
SANPO-Real-Train & ConvNeXt-L \cite{liu2022convnet} & 46.0 & 38.4\\

\midrule
\multicolumn{4}{c}{\textbf{Depth Estimation : BinsFormer \cite{li2022binsformer}}}\\
\midrule
\textbf{Dataset} & \textbf{Encoder}
& \multicolumn{2}{c}{\depthmetric{}$\uparrow$}\\
\midrule
SANPO-Synthetic-Train &  Resnet-50 \cite{he2016deep} & \multicolumn{2}{c}{0.27}\\
SANPO-Real-Train &  Resnet-50 \cite{he2016deep} & \multicolumn{2}{c}{0.45}\\

\bottomrule
\end{tabular}
}
\caption{\textbf{SANPO Baseline Performance}. Performance of panoptic segmentation and depth estimation methods trained on SANPO-Real-Train and SANPO-Synthetic-Train datasets.}
\label{table:train-baseline}
\vspace{-10pt}
}
\end{table}
We establish training baselines on SANPO using state-of-the-art architectures: Kmax-Deeplab \cite{yu2023kmaxdeeplab} for panoptic segmentation and BinsFormer \cite{li2022binsformer} for depth estimation. Table~\ref{table:train-baseline} presents the baseline performance of these models, trained on SANPO-Train and evaluated on SANPO-Real-Test. Only the human annotated subset of SANPO-Real-Test was used for evaluating panoptic segmentation.

\paragraph{Panoptic Segmentation} The results on SANPO (mIoU ranging from 40.0 to 48.1) fall significantly below those achieved on Cityscapes (mIoU ranging from 79.7 to 83.5). This disparity underscores the increased difficulty of the SANPO dataset, likely attributable to its substantially larger scale and greater diversity.

\paragraph{Depth Estimation} The BinsFormer architecture demonstrates a notable performance gap between KITTI and SANPO-Real. While achieving an impressive \depthmetric{} of 0.96 on KITTI \cite{li2022binsformer, Geiger2012CVPR}, the same architecture—BinsFormer with a Resnet-50 backbone—only reaches a \depthmetric{} of ~0.45 on SANPO-Real. This significant difference emphasizes the value of SANPO as a challenging benchmark for driving progress in depth estimation research.

\begin{table}[t]
{

\centering
\scalebox{0.9}{
\begin{tabular}{cccc}
\toprule
\multicolumn{4}{c}{\textbf{Panoptic Segmentation : Kmax-DeepLab \cite{yu2023kmaxdeeplab}}}\\
\midrule

\textbf{Dataset} & \textbf{Encoder} 
& \textbf{mIoU $\uparrow$}
& \textbf{PQ $\uparrow$}\\
\midrule

SANPO-Real-Train & Resnet-50 \cite{he2016deep} & 43.4 & 34.6\\

SANPO-Real-Train $\ddagger$ & Resnet-50 \cite{he2016deep} & 42.7 & 34.6\\


SANPO-Real-Train & ConvNeXt-L \cite{liu2022convnet} & 46.0 & 38.4\\
SANPO-Real-Train $\ddagger$  & ConvNeXt-L \cite{liu2022convnet} & 46.5 & 38.2\\

\bottomrule
\end{tabular}
}

\caption{\textbf{Accurate Machine Propagated Annotations}. Inclusion of machine propagated ground truth in the training of panoptic segmentation models does not compromise model performance.\\ $\ddagger$: Human annotated GT only.}
\label{table:train-machine-propagated}
}

\end{table}

We also investigated two key aspects of the SANPO dataset for panoptic segmentation:

\paragraph{Machine-Propagated Annotations} Table~\ref{table:train-machine-propagated} shows that restricting training to human-annotated ground truth has minimal impact on performance, validating the accuracy of SANPO's machine-propagated annotations. See~\ref{appendix:aot-accuracy} in the supplementary material
for an additional detailed quantitative analysis.

\begin{table*}[t]
{

\begin{center}
\begin{tabular}{cccc}
\toprule

\textbf{Dataset} & \textbf{Encoder} 
& \textbf{mIoU $\uparrow$}
& \textbf{PQ $\uparrow$}\\
\midrule

SANPO-Synthetic-Train & Resnet-50 \cite{he2016deep}	& 14.6	& 9.6\\
\begin{tabular}{@{}c@{}}SANPO-Synthetic-Train \textcolor{green}{\textbf{->}} SANPO-Real-Train$\ddagger$ \end{tabular} & Resnet-50 \cite{he2016deep} & 44.1 & 34.6\\
\begin{tabular}{@{}c@{}}SANPO-Synthetic-Train \textcolor{blue}{\textbf{+}} SANPO-Real-Train \\ \textit{80\% Synthetic} \dag\end{tabular} & Resnet-50 \cite{he2016deep} & 40.7 & 31.2\\
\begin{tabular}{@{}c@{}}SANPO-Synthetic-Train \textcolor{blue}{\textbf{+}} SANPO-Real-Train \\ \textit{50\% Synthetic} \dag\end{tabular} & Resnet-50 \cite{he2016deep} &	43.0 & 34.6\\
\midrule
SANPO-Synthetic-Train & ConvNeXt-L \cite{liu2022convnet} & 17.6	& 11.1\\
\begin{tabular}{@{}c@{}}SANPO-Synthetic-Train \textcolor{green}{\textbf{->}} SANPO-Real-Train $\ddagger$ \end{tabular} & ConvNeXt-L \cite{liu2022convnet} & 47.0 & 38.3\\
\begin{tabular}{@{}c@{}}SANPO-Synthetic-Train \textcolor{blue}{\textbf{+}} SANPO-Real-Train \\ \textit{80\% Synthetic} \dag\end{tabular} & ConvNeXt-L \cite{liu2022convnet} & 45.2 & 35.4\\
\begin{tabular}{@{}c@{}}SANPO-Synthetic-Train \textcolor{blue}{\textbf{+}} SANPO-Real-Train \\ \textit{50\% Synthetic} \dag \end{tabular} & ConvNeXt-L \cite{liu2022convnet} & \textbf{48.1} & \textbf{38.3}\\

\bottomrule
\end{tabular}
\end{center}
\caption{\textbf{Synthetic Data Boosts Performance}. Incorporating synthetic data, either through pretraining or as a component of the training dataset, consistently improves performance. Kmax-DeepLab \cite{yu2023kmaxdeeplab} is the method for all the rows. \textcolor{green}{\textbf{->}}: Pretrained on synthetic and then fine-tuned on real data. \textcolor{blue}{\textbf{+}} : Trained on combined real and synthetic data. $\ddagger$: Human annotated GT only.}
\label{table:train-synthetic-to-real}
}
\end{table*}

\paragraph{Synthetic to Real Domain Adaptation} Table~\ref{table:train-synthetic-to-real} demonstrates consistent improvements when pretraining on synthetic data before fine-tuning on real data (see rows marked with \text{->}). Training on both synthetic and real data also yields better results, although the optimal mixture varies.

    \section{Applications}
SANPO serves as the flagship training set for Project Guideline \cite{PROJECTGUIDELINE}, a mobile application which uses SANPO-trained mobile-friendly depth estimation and semantic segmentation models to help people with low vision walk and run independently. The Project Guideline authors have already open-sourced their models trained on SANPO.\cite{PROJECTGUIDELINE}

\paragraph{Obstacle detection task}
SANPO focuses on depth estimation and panoptic segmentation because these are critical building blocks required by outdoor visual navigation tasks.
This makes SANPO uniquely suited for these downstream navigation applications.
To showcase this, we designed a simple on-device obstacle localization benchmark using SANPO-Real’s depth and panoptic labels.
In this benchmark, we focus on three capabilities: (i) detecting walkable vs non-walkable surfaces, (ii) recognizing and detecting obstacles, (iii) determining ego person's distance to the obstacles. These tasks are appropriate because tasks (ii) and (iii) need both segmentation and depth estimation, while (i) mainly requires reliable segmentation.
\textbf{Metrics.} For this task, we mapped SANPO labels to ``safe for walking,'' ``not safe for walking,'' and ``obstacles'' (Refer to \ref{appendix:accessibility-mapping} in the supplementary material). For (i) and (ii), we report mIoU. For (iii), we report 
<=1.25 of only obstacle classes, binning the metric at short range, medium range, and long range.
\textbf{Baselines.} For semantic segmentation, we trained a quantized MobileNetV3\cite{howard2019searching} with an input resolution of 512$\times$512. For depth estimation, we used a small MidasNet model~\cite{DBLP:journals/corr/abs-1907-01341} with an input resolution of 192$\times$192. Depth output is resized to 512$\times$512, with metrics reported at this resolution. These lightweight models run in real-time at 30FPS on Pixel 6 phones.
See Table~\ref{table:accessibility-eval} for results.

\begin{table}[t]
\centering
\begin{tabular}{ccc}
\toprule
 \multicolumn{3}{c}{\textbf{Segmentation mIoU $\uparrow$}}\\
 \midrule
 safe for walking & not safe for walking & obstacles \\
 83.4 & 64.5 & 85.9 \\
\midrule
 \multicolumn{3}{c}{\textbf{Obstacles' Depth Estimation $\depthmetric~\uparrow$}}\\
 \midrule
<10 meters & < 15 meters & <80 meters\\
 0.27 & 0.25 & 0.20\\
 \bottomrule
\end{tabular}

%
%
\caption{\textbf{On-device Navigation specific Segmentation and Obstacle Detection.}}
\label{table:accessibility-eval}
\vspace{-10pt}
\end{table}


    \section{Limitations and Future Work}


SANPO focuses on outdoor environments, but it does not cover all possible navigation scenarios for the visually impaired. Limitations of SANPO include USA-centric geography, lack of indoor environments, and lack of nighttime data.
Most comparable scene understanding datasets
(KITTI \cite{Geiger2012CVPR}, SideGuide\cite{9340734}, NYU \cite{Silberman:ECCV12}, SCAND \cite{karnan2022socially}, MuSoHu \cite{nguyen2023toward}, etc.)
are collected in a single geographic location. In comparison, SANPO, similar to Waymo Open \cite{mei2022waymo}, was collected 
in four regions across the USA: California (San Francisco and Mountain View), Colorado (Boulder), and New York City. This represents a geographically diverse blend of West Coast, Rocky Mountains, and Northeast environments in urban, suburban, and park settings.
We chose these four regions in a single country because collecting data like SANPO is a very involved, costly, and manpower-intensive process and has to pass several state, city and local legal requirements.
We excluded indoor environments because of legal and privacy considerations. 
For low-light and nighttime data, 10\% of sessions in SANPO-Synthetic are rendered from nighttime environments. We kept this at 10\% to prevent the domain similarity from diverging between SANPO-Real and SANPO-Synthetic. 
Extending our efforts to other countries, regions, and indoor environments, as well as including more nighttime data is something we will consider in the future. Additional synthetic-to-real experiments is future work as well.

    \section{Conclusion}
This paper introduces SANPO, a large-scale egocentric video dataset that accelerates the development of computer vision-based assistive technologies. SANPO provides a comprehensive resource for researchers, offering diverse real and synthetic data alongside mobile-friendly, pre-trained models for semantic segmentation and depth estimation.  
We open-sourced both the dataset and models under the CC BY 4.0 license\footnote{http://creativecommons.org/licenses/by/4.0} to empower the CV community in building robust human egocentric navigation systems.

    {\small
    \bibliographystyle{ieee_fullname}
    \bibliography{main}
    }

   \clearpage
   
  \fi 

\fi 

\end{document}